\documentclass[twocolumn]{svjour3}         
\smartqed  
\usepackage{graphicx}

\usepackage{titlesec} 
\titleformat{\paragraph}[runin]
{\bfseries\scshape}{\theparagraph}{1em}{}
\titlespacing{\paragraph}{0em}{1ex}{.5em} 
\usepackage{todonotes}
\usepackage{booktabs}
\usepackage{tabularx}
\usepackage{multirow}
\usepackage{multicol}
\usepackage{bm}
\usepackage{url}
\usepackage{balance}
\usepackage{enumitem}
\usepackage{color}
\usepackage{bbding}
\usepackage{nccmath}
\usepackage{amsmath,amssymb}
\usepackage{graphicx}
\usepackage{cite}

\usepackage{graphicx}
\usepackage{amsmath}
\usepackage{amssymb}
\usepackage{color}
\usepackage{epstopdf}
\usepackage{array}
\usepackage{multirow}
\usepackage{amsfonts}
\usepackage{booktabs}
\usepackage{float}
\usepackage{subfigure}
\usepackage{algorithm}
\usepackage{algpseudocode}
\usepackage[switch]{lineno}

\begin{document}

\title{Heterogeneous Semantic Transfer for Multi-label Recognition with Partial Labels
}


\author{Tianshui Chen \and
        Tao Pu \and
        Lingbo Liu \and
        Yukai Shi \and
        Zhijing Yang \and
        Liang Lin
}


\institute{Tianshui Chen \at
             Guangdong University of Technology, Guangzhou, China \\
              \email{chentianshui@gmail.com}
           \and
           Tao Pu \at
           Sun Yat-sen University, Guangzhou, China \\
              \email{putao537@gmail.com}
           \and
          Lingbo Liu \at
            The Hong Kong Polytechnic University \\
              \email{lingbo.liu@polyu.edu.hk}
           \and
          Yukai Shi \at
           Guangdong University of Technology, Guangzhou, China \\
              \email{ykshi@gdut.edu.cn}
           \and
          Zhijing Yang \at
           Guangdong University of Technology, Guangzhou, China \\
              \email{yzhj@gdut.edu.cn}
           \and
          Liang Lin \at
           Sun Yat-sen University, Guangzhou, China \\
              \email{linlng@mail.sysu.edu.cn}
}

\date{Received: date / Accepted: date}

\maketitle


\def\eg{\emph{e.g.}}
\def\etal{\emph{et al}}

\newcommand{\EQREF}{Eq.~\eqref}
\newcommand{\EQSREF}{Eqs.~\eqref}
\newcommand{\FIGREF}{Fig.~\ref}
\def\proposed{VB} 
\def\fixcolor{black}
\def\hstate{\bm {\tilde s}}
\def\rstate{\bm s}
\def\jstate{\bm s^{jn}}
\def\jpolicy{\overrightarrow{\pi}}
\def\vpref{v_{\text {pref}}}
\def\vector#1{\mbox{\boldmath $#1$}}
\def\sup#1{^{(\rm #1)}}
\def\sub#1{_{\rm #1}}
\def\supi#1{^{(#1)}}
\def\vct#1{\mbox{\boldmath $#1$}}
\def\eg{{\it e.g.}}
\def\cf{{\it c.f.}}
\def\ie{{\it i.e.}}
\def\etal{{\it et al. }}
\def\etc{{\it etc}}
\newcommand{\argmax}{\mathop{\rm argmax}\limits}
\newcommand{\argmin}{\mathop{\rm argmin}\limits}

\def\Rerr{\Delta \bm r}
\def\Terr{\Delta \bm t}
\def\Xerr{\Delta \bm x}
\def\XerrRel{\Delta \bm {\tilde x}}
\def\Xgt{\dot{\bm x}}
\def\Rgt{\dot{R}}
\def\Tgt{\dot{\bm t}}
\def\arraystretchlen{1.0}

\def\cam{c}
\def\image{\mathcal I}
\def\traj{\mathcal X}
\def\btraj{\mathcal {\bm X}}
\def\keypoints{\mathcal P}
\def\states{\mathcal S}
\def\bstates{\mathcal {\bm S}}
\def\state{\bm s}
\def\ped{\bm x}
\def\pedi{\bm p} 
\def\obs{\bm z}

\def\Fi{\bm F_r}
\def\Fp{\bm F_p}
\def\vpref{\bm w}
\def\ENERGY{{\mathcal E}}

\def\DIFF#1{\textcolor{black}{#1}}
\def\DIFFCR#1{\textcolor{black}{#1}}

\begin{abstract}
Multi-label image recognition with partial labels (MLR-PL), in which some labels are known while others are unknown for each image, may greatly reduce the cost of annotation and thus facilitate large-scale MLR. We find that strong semantic correlations exist within each image and across different images, and these correlations can help transfer the knowledge possessed by the known labels to retrieve the unknown labels and thus improve the performance of the MLR-PL task. In this work, we propose a novel heterogeneous semantic transfer (HST) framework that consists of two complementary transfer modules that explore both within-image and cross-image semantic correlations to transfer the knowledge possessed by known labels to generate pseudo labels for the unknown labels. Specifically, an intra-image semantic transfer (IST) module learns an image-specific label co-occurrence matrix for each image and maps the known labels to complement the unknown labels based on these matrices. Additionally, a cross-image transfer (CST) module learns category-specific feature-prototype similarities and then helps complement the unknown labels that have high degrees of similarity with the corresponding prototypes. It is worthy-noting that the HST framework requires searching appropriate thresholds to determine the co-occurrence and similarity scores to generate pseudo labels for the IST and CST modules, respectively. To avoid highly time-consuming and resource-intensive manual tuning, we introduce a differential threshold learning algorithm that adjusts the nondifferential indication function to a differential formulation to automatically learn the appropriate thresholds. Finally, both the known and generated pseudo labels are used to train MLR models. Extensive experiments conducted on the Microsoft COCO, Visual Genome, and Pascal VOC 2007 datasets show that the proposed HST framework achieves superior performance to that of current state-of-the-art algorithms. Specifically, it obtains mean average precision (mAP) improvements of 1.4\%, 3.3\%, and 0.4\% on the three datasets over the results of the best-performing previously developed algorithm.
\end{abstract}

\section{Introduction} 
\label{intro}
\begin{figure}[!t]
   \centering
   \includegraphics[width=0.95\linewidth]{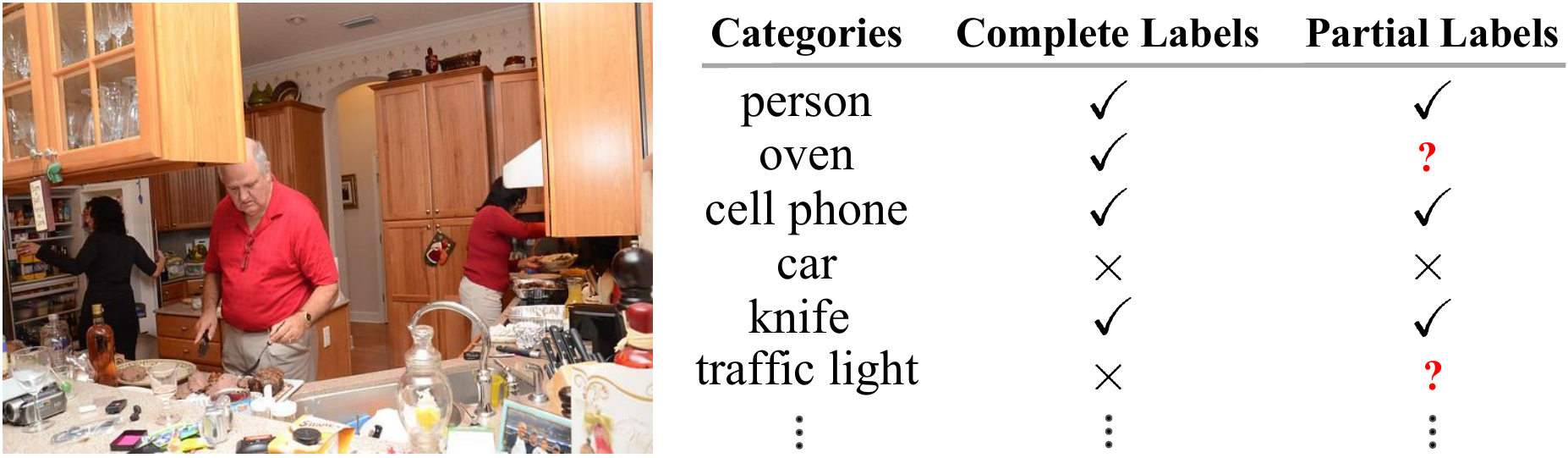}
   \caption{Illustration of MLR with complete and partial label settings. For the complete label setting, $\checkmark$ denotes that the corresponding category exists, while $\times$ denotes that it does not exist. For the partial label setting, $\textbf{?}$ denotes that the original label is missing and we do not know whether it exists or does not exist. Both the $\checkmark$ and $\times$ labels are regarded as known labels, while the $\textbf{?}$ labels are regarded as unknown labels.}
   \label{fig:task}
\end{figure}

Multi-label image recognition (MLR) is a more practical and necessary task than its single-label counterpart because images of daily life generally contain multiple objects belonging to diverse categories. Recently, many efforts \cite{chen2019multi,chen2019learning,chen2022knowledge, zhou2022learning} have been dedicated to addressing this task, as it benefits various applications ranging from content-based image retrieval \cite{li2010technique,zhang2021instance} and person re-identification \cite{wang2022unsupervised, lan2020semi, zhu2021intra} to recommendation systems \cite{carrillo2013multi,zheng2014context} and facial expression recognition \cite{li2019blended, ruan2022adaptive}. Despite the acknowledged progress, current leading algorithms \cite{chen2019multi,chen2019learning,chen2022knowledge} utilize notoriously data-hungry deep convolutional networks \cite{he2016deep,simonyan2015very} to learn discriminative features; such networks depend heavily on large-scale datasets that have clean and complete annotations. However, it is extremely time-consuming and labor-intensive to annotate a consistent and exhaustive list of labels for each image. In contrast, it is easy and scalable to annotate partial labels for each image, which can be regarded as an alternative way to address the above problem. In this work, we aim to address the task of learning MLR models with partial labels (MLR-PL), i.e., merely a few positive and negative labels are known, while the others are unknown, as shown in Figure \ref{fig:task}.

\begin{figure}[!t]
   \centering
   \includegraphics[width=0.9\linewidth]{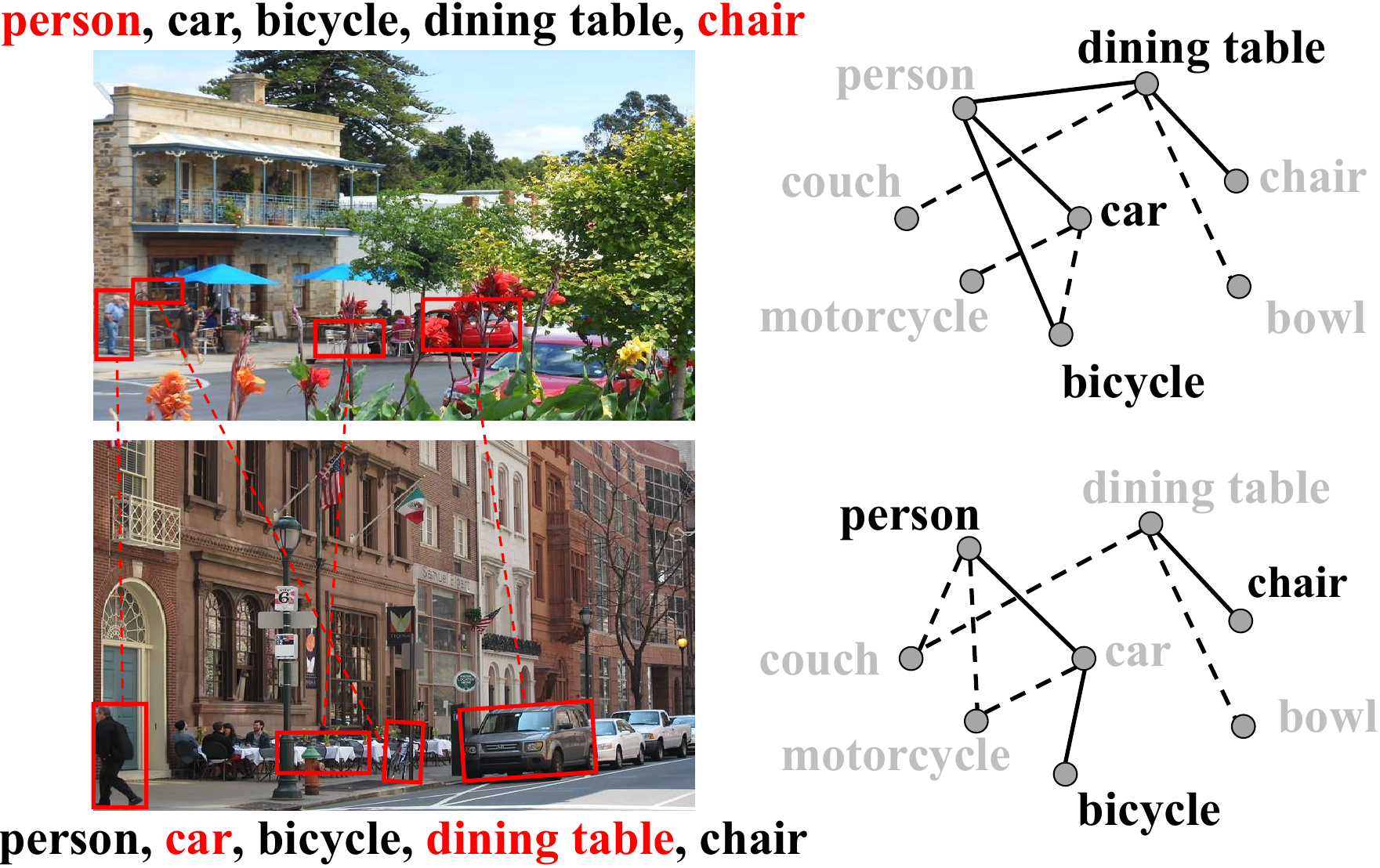}
   \caption{Two examples of images with partial labels (the unknown labels are highlighted in red). Strong semantic correlations exist with each image and across different images, and we can mine these correlations to help complement the unknown labels. }
   \label{fig:motivation}
\end{figure}

Current algorithms mainly formulate MLR as a multiple binary classification task. Regarding the unknown labels as missing or negative labels is an intuitive and simple way to adapt these algorithms to address the MLR-PL task \cite{sun2017revisiting,joulin2016learning}. However, such an adaptation loses some data or even incurs some noisy labels, which may hurt the training process and inevitably lead to an obvious performance drop. Fortunately, strong semantic correlations exist within each image and across different images. These correlations can help to efficiently transfer the semantic knowledge of known labels to construct the unknown labels and thus solve the aforesaid dilemma. As shown in Figure \ref{fig:motivation}, correlations belong to two types, which are described as follows. i) Within-image correlations: Label co-occurrences are widespread in images of the real world, and labels with high co-occurrence probabilities may co-exist in one image, e.g., cars tend to co-occur with persons, while tables and chairs are likely to co-exist. ii) Cross-image correlations: Objects that belong to the same category and come from different images may share similar visual appearances, and thus images with similar visual features may have the same labels.

In this work, we explore the mining of these correlations to help complement the unknown labels via a novel heterogeneous semantic transfer (HST) framework. This framework consists of two complementary modules that learn image-specific co-occurrence matrices to help the transfer semantic labels within each image and the category-specific feature-prototype similarities to transfer semantic labels across different images. Although a previous work \cite{huynh2020interactive} also noticed label/image dependencies, it merely introduced a statistical co-occurrence matrix and image-level similarities to regularize the training process of MLR models. In contrast, the HST framework can learn image-specific cooccurrence matrices and category-specific feature-prototype similarities, which can capture finer-grained correlations to help construct accurate pseudo labels for the unknown labels to facilitate the MLR-PL task. For example, in Figure \ref{fig:motivation}, the feature vectors of the \emph{truck} are similar in two different images, and we can use the annotated \emph{truck} in the upper image to help complement the unknown \emph{truck} in the lower image. Similarly, the \emph{traffic light} has a high co-occurrence probability with the \emph{car}, and we can complete this unknown label based on the co-occurrence.

The HST framework builds on a semantic-aware representation learning (SARL) module, which incorporates category semantics to learn category-specific feature representations. Then, an intra-image semantic transfer (IST) module is designed to learn a co-occurrence matrix among all categories for each image and complement the unknown labels that have high co-occurrence probabilities with the known labels. Furthermore, a cross-image semantic transfer (CST) module is introduced to learn the category-specific representation prototypes and category-specific similarities between the corresponding feature representations and prototypes. It then transfers the semantics of the known labels to help complement some unknown labels with high similarity. Finally, the known labels and complemented labels are used to supervise the training of the MLR model. The HST training process, together with the online generation of pseudo labels, is seamlessly conducted in an end-to-end manner.

To generate pseudo labels, the IST module requires a threshold to determine the co-occurrence probability values, while the CST module depends on a threshold to determine the similarity value for each setting. Manually setting the threshold values via a cross-validation pipeline is an intuitive and simple method. However, searching the optimal values for different datasets and known label proportions is extremely time-consuming and resource-intensive. To address this dilemma, we introduce a differential threshold learning (DTL) algorithm that adjusts the nondifferential indication function to a differential formulation and thus can make full use of the known labels to automatically learn the appropriate thresholds. In this way, the algorithm can avoid the highly time-consuming and resource-intensive hyperparameter searching process, making the proposed HST framework more practical for real-world scenarios.

A preliminary version of this work was presented as a conference paper \cite{chen2021structured}. In this version, we inherit the idea of structured semantic transfer and strengthen this work in terms of the following four aspects. First, instead of learning instance-level feature similarities, the CST module is designed to learn category-specific representation prototypes and feature-prototype similarities that can help generate more robust and accurate pseudo labels. The detailed descriptions of this module are presented in Sec. \ref{sec:cst}, and the performance comparisons with the instance-level counterpart are presented in the third paragraph of Sec. \ref{sec:exp-cst}. Second, a DTL algorithm is introduced to adaptively search the optimal threshold values for different datasets and settings, avoiding the highly time-consuming and resource-intensive hyperparameter tuning process. We present the detailed introduction to this algorithm in Sec. \ref{sec:dtl} and verify its effectiveness in Sec. \ref{sec:exp-dtl}. Third, we introduce different algorithms to implement the SARL module, demonstrating the feasibility of the proposed framework. We add the two implementation algorithms in Sec. \ref{sec:sarl} and their performance comparisons in Sec. \ref{sec:exp-sarl}. Finally, substantially more experiments are conducted to verify the effectiveness of the proposed framework and evaluate the actual contribution of each crucial module for a better understanding of the HST framework. Specifically, by improving the CST module and introducing the DTL algorithm, it improves the average mAP by 1.3\%, 3.0\%, and 0.5\% on the MS-COCO, VG-200, and Pascal VOC 2007 datasets compared with the conference version.

The contributions of this work are summarized into four folds.

\begin{itemize}

\item We introduce an HST framework to simultaneously mine intra-image and cross-image correlations to help complement the unknown labels and thus facilitate the MLR-PL task.

\item We propose an IST module to learn image-specific co-occurrences and help generate pseudo labels and a CST module to learn category-specific representation prototypes and feature-prototype similarities to help complement the unknown labels.

\item We design a DTL algorithm that can make use of the known labels to automatically learn the thresholds for different datasets and settings.

\item We conduct extensive experiments on variant datasets (e.g., Microsoft COCO, Pascal VOC, and Visual Genome) to demonstrate the effectiveness of the proposed HST framework. We also perform ablation studies to analyze the contribution of each module to attain a better understanding of our framework. Trained models and codes are available at \textbf{\url{https://github.com/HCPLab-SYSU/HCP-MLR-PL}}.

\end{itemize}

The rest of this paper is organized as follows. We review the works that are related to MLR with complete and partial labels in Section \ref{sec:related_work}. We then introduce the proposed HST framework in Section \ref{sec:HST} and present the experiments and evaluations in Section \ref{sec:exp}. Finally, we conclude the work in Section \ref{sec:conclusion}.

\section{Related Works}
\label{sec:related_work}
MLR has received increasing attention \cite{wei2016hcp,yang2016exploit,chen2019multi,chen2019learning,nie2022multi,chen2022knowledge} since it is more practical and necessary than its single-label counterpart. To solve this task, many efforts have been dedicated to discovering discriminative local regions for feature enhancement via object proposal algorithms \cite{wei2016hcp,yang2016exploit} or visual attention mechanisms \cite{ba2014multiple,chen2018recurrent,wang2017multi}. Another line of works proposed capturing label dependencies to regularize the training of MLR models and thus improve their performance \cite{wang2016cnn,wang2017multi,chen2019multi,chen2019learning,Pu2024STKET}. These works either introduced a recurrent neural network (RNN)/long short-term memory (LSTM) \cite{hochreiter1997long,liu2019contextualized} to implicitly capture label dependencies \cite{wang2016cnn,wang2017multi} or explicitly modeled the label dependencies in the form of a structured graph and exploited graph neural networks \cite{li2016gated,chen2022cross} to adaptively capture label dependencies. Recently, Chen et al. \cite{chen2019learning} presented state-of-the-art results obtained on several multi-label datasets by using semantic-specific graph representation learning (SSGRL) to obtain semantic-aware features for different category labels, and we employ their SSGRL module for learning category-specific features in this work. However, despite achieving remarkable progress, all these methods rely on data-hungry deep neural networks (DNNs) \cite{simonyan2015very,he2016deep} to learn discriminative feature representations and thus require large-scale and clean datasets (e.g., Visual Genome \cite{krishna2017visual}, MS-COCO \cite{lin2014microsoft} and Pascal VOC \cite{everingham2010pascal}) for training. However, annotating a complete list of labels for every image is time-consuming and labor-intensive, making the collection of large-scale and complete multi-label datasets less practical and scalable.

An alternative way is learning MLR models with partial labels; i.e., only some of the labels are known while the rest labels are unknown \cite{durand2019learning,huynh2020interactive,sun2017revisiting,joulin2016learning,tsoumakas2007multi}. Humans can quickly capture some objects at a glance, but require much more time to annotate a consistent and exhaustive list of labels for each image. Thus, learning MLR models with partial labels can help reduce the annotation cost and thus make MLR more scalable and practical. To address this task, some works \cite{bucak2011multi,wang2014binary,sun2010multi,sun2017revisiting} have simply regarded the unknown labels as negative labels and trained the models with a similar scheme for the fully labeled setting. These methods may suffer from severe performance drops, as many positive labels are wrongly annotated as negative, which has an obvious impact on model training. Other works \cite{tsoumakas2007multi} have treated MLR as a task involving multiple independent binary classifications. However, this approach ignores the label dependencies that play a key role in MLR. To overcome this issue, some other works have exploited label dependencies to transfer the known labels to help complement the unknown labels \cite{xu2013speedup,cabral2011matrix,yu2014large,wu2015ml,KapoorVJ12nips}. Cabral et al. \cite{cabral2011matrix} introduced low-rank regularization to exploit label correlations and complete unprovided labels, while Wu et al. \cite{wu2015ml} similarly adopted low-rank empirical risk minimization. A mixed graph was utilized in \cite{WuLG15iccv} to encode a network of label dependencies. In \cite{KapoorVJ12nips}, missing labels were treated as latent variables in probabilistic models and predicted by posterior inference through Bayesian networks. However, most of these works depend on solving an optimization problem that requires loading the whole training set into memory, and they cannot be incorporated into DNNs \cite{simonyan2015very,he2016deep} that are trained via the minibatch strategy. Such limitations result in inferior performance since fine-tuning is critical for transferring pretrained DNN models. More recently, Durand et al. \cite{durand2019learning} proposed a loss that generalizes the standard binary cross-entropy (BCE) loss by exploiting label proportion information and uses it to train a DNN \cite{he2016deep} with partial labels. Huynh et al. \cite{huynh2020interactive} introduced statistical co-occurrences and image-level feature similarity to regularize the standard BCE loss for DNNs. Pu et al. \cite{pu2022semantic} aim to blend category-specific feature from different images or prototypes and thus generate blended features with combined labels. This process can be regarded as augmenting new samples in feature spaces.

Different from these methods, the proposed framework introduces two complementary modules, where the first module learns image-specific co-occurrence correlations to transfer the provided labels within the same image to complement unknown labels, and the second module learns category-level representation prototypes and feature-prototype similarity correlations to transfer the provided labels across different images to complement unknown labels. The two modules can be effortlessly integrated into current MLR models, enabling the generation of pseudo labels in an online manner. This integration allows for the simultaneous execution of the MLR model training and pseudo label generation processes, ensuring a cohesive and efficient pipeline.

\begin{figure*}[!t]
   \centering
   \includegraphics[width=0.9\linewidth]{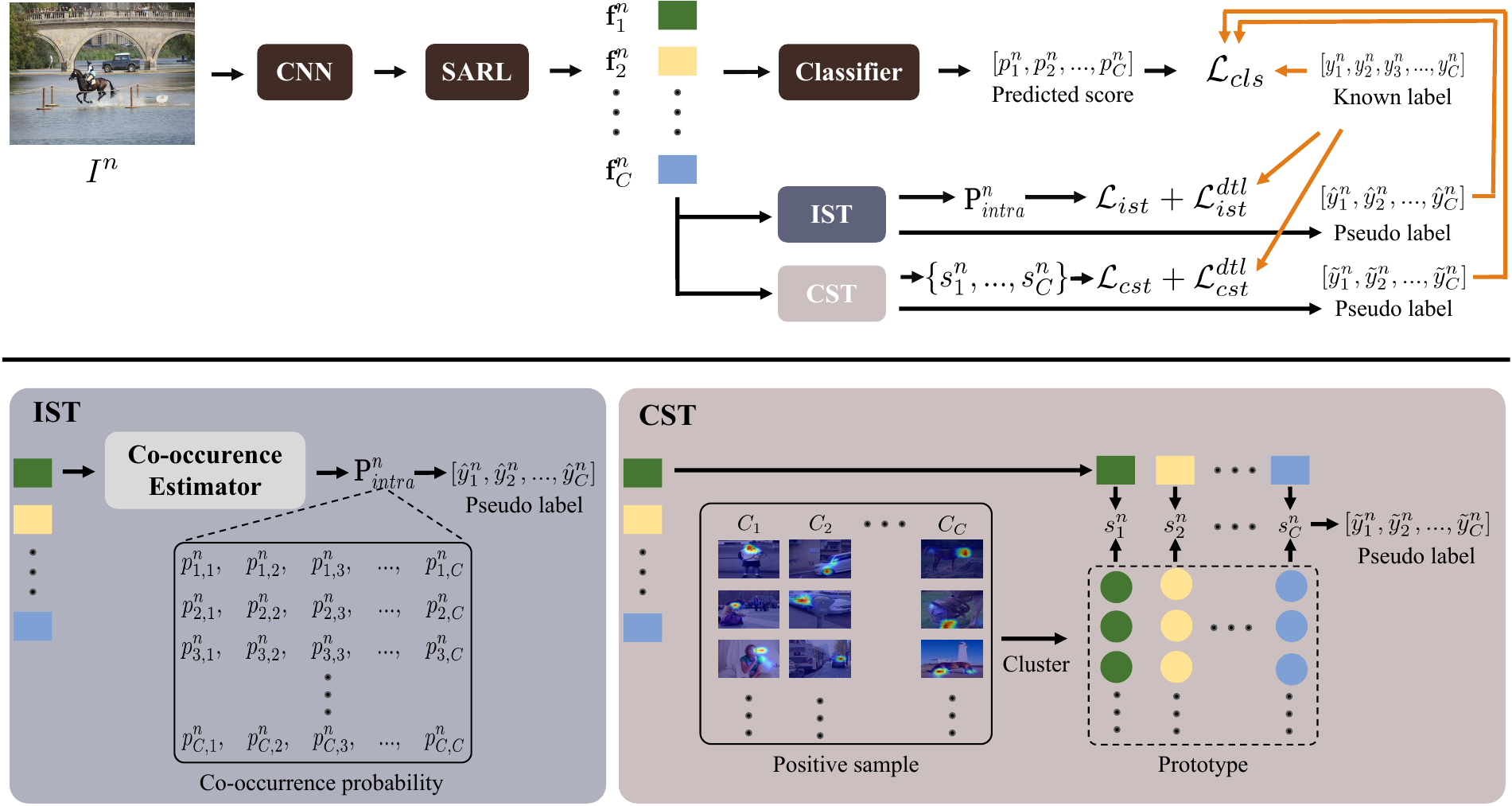}
   \caption{An overall illustration of the proposed HST framework. The upper part is the overall pipeline that consists of the IST and CST modules for generating pseudo labels, which are then fed to the supervised training process of the MLR model. The lower part contains the detailed implementations of the IST and CST modules. The IST module first predicts label co-occurrence matrices and then maps the known labels to complement the unknown labels. The CST module first learns category-specific feature and prototype similarities across different images and then maps them to generate pseudo labels.}
   \label{fig:framework}
\end{figure*}

\section{HST}
\label{sec:HST}
In this section, we introduce the proposed HST framework that mines intra-image and cross-image correlations to help complement the unknown labels. It builds on a SARL module that extracts category-specific feature vectors for each image. The IST module first learns the co-occurrence probability of each category pair and then constructs a label co-occurrence matrix for each image. It then generates pseudo labels for the unknown labels that have high co-occurrence probabilities with the known labels. In this way, it transfers the semantic knowledge of the known labels to complement the unknown labels. Furthermore, the CST module learns the category-specific representation prototype and feature-prototype similarities. Then, it generates pseudo labels for the unknown labels that have high similarities with the prototypes of the corresponding labels. Finally, we can obtain accurate pseudo labels for the unknown labels and use both the generated pseudo labels and the known labels to train the multi-label model. An overall illustration is presented in Figure \ref{fig:framework}.

\noindent\textbf{Notation. }Here, we introduce the notations used throughout the paper. We denote the training dataset as $\mathcal{D}=\{(I^1,y^1), ..., (I^N,y^N)\}$, in which $N$ is the number of training samples. $\textbf{y}^n=\{y^n_1, y^n_2, \cdots, y^n_C\}\in \{-1, 0, 1\}^C$ is the label vector for the $n$-th sample, and $C$ is the number of label. $y^n_c$ is assigned to 1 if label $c$ exists in the $n$-th image, assigned to -1 if it does not exist, and assigned to 0 if it is unknown.

\subsection{SARL}
\label{sec:sarl}
The SARL module is introduced to incorporate category information to extract category-specific feature vectors for images. Given an input image $I^n$, we first utilize a backbone network to extract a pool of global feature maps $\textbf{f}^n$ and then utilize the SARL module to extract a feature vector for each category. This operation is formulated as
\begin{equation}
 [\textbf{f}^n_1, \textbf{f}^n_2, \cdots, \textbf{f}^n_C]=\phi_{sarl}(\textbf{f}^n).
\end{equation}
Here, we utilize current algorithms to implement this module, i.e., semantic decoupling (SD) \cite{chen2019learning} and semantic attention (SA) \cite{Ye2020ADD-GCN}, as described in the following.

\noindent\textbf{SD} first extracts a semantic embedding vector for each category and then incorporates this vector to guide the process of learning feature vectors for different categories that pay more attention to the corresponding regions.

\noindent\textbf{SA} first generates category-specific activation maps by using class activation mapping \cite{Zhou2016CAM} and then utilizes these maps to convert the transformed feature map into a content-aware vector for each category.

\noindent Once we obtain the category-specific feature vectors, according to previous works \cite{chen2019learning,Ye2020ADD-GCN}, we adopt a gated graph neural network followed by fully connected layers to compute the confidence score for each category. We finally use a sigmoid function on each score to compute the corresponding probability vector:
\begin{equation}
 \centering
  [p^n_1, p^n_2, \cdots, p^n_C]=\phi([\textbf{f}^n_1, \textbf{f}^n_2, \cdots, \textbf{f}^n_C]).
\end{equation}

It is crucial that the category-specific feature vector $\textbf{f}^n_i$ can learn information that is actually relevant to category $i$. Both works \cite{chen2019learning,Ye2020ADD-GCN} ensure this point via two mechanisms. First, they introduce semantic information as guidance to learn feature vectors for the corresponding categories. Concretely, SD \cite{chen2019learning} exploits semantic representation (i.e., Glove features \cite{pennington2014glove}) and SA \cite{Ye2020ADD-GCN} exploits the class activation maps to incorporate semantic information to guide class-relevant features for each class. Second, it mainly depends on $\textbf{f}^n_i$ to predict whether the category $i$ exists, which can also prompt learning category-specific features.

\subsection{IST}
Strong co-occurrence correlations are present among the semantic labels in real-world images, and naturally, these correlations can effectively guide the transfer of semantic knowledge derived from known labels to generate pseudo labels for the unknown labels. A recent work \cite{huynh2020interactive} applied dataset-level statistical correlations to achieve this goal. However, statistical correlations are not appropriate for every image and thus inevitably incur some incorrect labels. To avoid this problem, the IST module is proposed to learn an image-specific label co-occurrence matrix and apply this matrix to complement the unknown labels for the corresponding image.

Given the feature vectors $[\textbf{f}^n_1, \textbf{f}^n_2, \cdots, \textbf{f}^n_C]$ for an input image $I^n$, we need to compute the co-occurrence probability for each category pair. For categories $i$ and $j$, we first concatenate their feature vectors $\textbf{f}^n_i$ and $\textbf{f}^n_j$ and then feed the concatenated features to compute their co-occurrence probability, which is formulated as
\begin{equation}
 p_{i,j}^{n}=\phi_{intra}([\textbf{f}^n_i, \textbf{f}^n_j]),
\end{equation}
where $\phi_{intra}(\cdot)$ is implemented by several stacked fully connected layers. We compute the probabilities for all pairs and obtain a co-occurrence matrix $P_n\in \mathcal{R}^{C\times C}$. Then, we estimate the pseudo labels for the unknown labels based on the co-occurrence matrix and the known labels. For a category $i$ that is not provided, we can compute its pseudo label by
\begin{equation}
 \hat{y}^{n, intra}_{i}=\textbf{1}[(\sum_{\{j|y^{n}_{j}=1\}}p_{i,j}^{n} \cdot y^{n}_{j}) \ge \theta_{intra}],
 \label{equ:intra-label}
\end{equation}
where $\textbf{1}[\cdot]$ is an indicator function whose value is 1 if the argument is positive and is 0 otherwise. $\theta_{intra}$ is a threshold that helps to exclude the unlikely labels. We compute the pseudo labels for all unknown labels and set the known labels to 0, obtaining $\textbf{y}^{n, intra}=\{y^{n, intra}_1, y^{n, intra}_2, \cdots, y^{n, intra}_C\}$.

Formally, the co-occurrence prediction problem can be considered a binary classification task, and we can train the associated model by using the BCE loss. However, it is very difficult to train the co-occurrence predictor because the positive and negative pairs are extremely imbalanced. To address this task, we introduce the asymmetric loss \cite{ridnik2021asymmetric} that dynamically downweights the importance of easy negative pairs; the loss function is defined as
\begin{equation}
\mathcal{L}_{ist}=\sum_{n=1}^N \sum_{\{i,j\}}\ell_{i,j}^n,
\end{equation}
where
\begin{equation}
\ell_{i,j}^n=
    \begin{cases}
         (1-p_{i,j}^n)^{\gamma_1}\log(p_{i,j}^n) \quad &\{i,j\}\in \mathcal{D}^n\\
         (p_{i,j}^n-m)^{\gamma_2}\log(1-p_{i,j}^n)  \quad &\{i,j\}\notin \mathcal{D}^n.
    \end{cases}
\end{equation}
Here, $\mathcal{D}^n$ is the set of label pairs that co-occur in image $n$. $\gamma_1$, $\gamma_2$, and $m$ are three parameters used to balance the loss, and they have been fully explored in work \cite{ridnik2021asymmetric}. Thus, we follow this work to set them to 1, 2, and 0.05, respectively.

\subsection{CST}
\label{sec:cst}
Intuitively, objects belonging to the same category but present in different images share similar visual appearances. In other words, if two images share similar visual features, they tend to have the same labels. In the context of multi-label images, two difficulties are faced when taking advantage of these correlations. First, each image contains multiple objects that belong to diverse categories, and image-level similarities can hardly measure object similarities. Second, objects that belong to the same category but are derived from different images may share diverse appearances, and instance-level similarities cannot accurately measure label correlations. In this work, we design the CST module to learn representation prototypes for each category, estimate category-specific feature-prototype similarities and complement the unknown labels that have high similarities with the corresponding prototypes.

The prototypes are used to represent the overall descriptions of the corresponding categories. For each category $c$, we first select all the images that are annotated with label $c$ and extract the feature vectors of label $c$ for all these images. This produces a set of feature vectors, i.e., $[\textbf{f}^{1}_c, \textbf{f}^{2}_c, \cdots, \textbf{f}^{N_c}_c]$, in which $N_c$ denotes the number of selected images. Then, the simple K-means algorithm is used to cluster these features into $K$ prototypes, i.e., $P_{c}=[\textbf{p}^1_c, \textbf{p}^2_c, ..., \textbf{p}^K_c]$. We repeat the above process to extract the prototypes for all categories.

For each category $c$ of images $I^n$, we compute the cosine distance between the feature vector $\mathbf{f}_c^n$ and the prototype $\mathbf{p}_c^k$, which is formulated as
\begin{equation}
 s^{n, k}_c=cosine(\textbf{f}^n_c, \textbf{p}^k_c) = \frac{\textbf{f}^n_c\cdot\textbf{p}^k_c}{||\textbf{f}^n_c||\cdot||\textbf{p}^k_c||}.
\end{equation}

Given an unknown label $c$ in image $n$, we can compute the similarities between $\textbf{f}^n_c$ and the corresponding prototypes $[\textbf{p}^1_c, \textbf{p}^2_c, ..., \textbf{p}^K_c]$. We average all the similarities to obtain its overall similarity with category c (i.e., $s_c^n$) and estimate the existence of category c by

\begin{equation}
  \hat{y}^{n, cross}_{c}=\textbf{1}[(s^n,c \ge \theta_{cross}].
 \label{equ:cross-label}
\end{equation}
Similarly, $\textbf{1}[\cdot]$ is an indicator function, and $\theta_{cross}$ is a threshold. We also estimate the pseudo labels for all unknown labels and set the known labels to 0, obtaining $\textbf{y}^{n, cross}=\{y^{n, cross}_1, y^{n, cross}_2, \cdots, y^{n, cross}_C\}$.

To obtain compact representation prototypes, it is expected that the feature vectors that belong to the same category have high similarities. To achieve this end, the similarity between $\textbf{f}^m_c$ and $\textbf{f}^n_c$ should be high if images $m$ and $n$ have the same known label $c$, and the similarity should be low otherwise. For a category $c$ of images $m$ and $n$, we define instance-level similarity by using the cosine distance; this is formulated as follows:
\begin{equation}
 s^{n,m}_c=cosine(\textbf{f}^n_c, \textbf{f}^m_c) = \frac{\textbf{f}^n_c\cdot\textbf{f}^m_c}{||\textbf{f}^n_c||\cdot||\textbf{f}^m_c||}.
\end{equation}
Then, we formulate the above objective as a ranking task and introduce a paired loss for training; this loss function is formulated as
\begin{equation}
\mathcal{L}_{cst}=\sum_{n=1}^N\sum_{c=1}^C\sum_{m} \ell^{m,n}_c,
\end{equation}
where
\begin{equation}
\ell^{m,n}_c=
    \begin{cases}
         1-s^{m,n}_c \quad & y_c^m=1,y_c^n=1\\
         1+s^{m,n}_c \quad & otherwise.
    \end{cases}
\end{equation}

\subsection{DTL}
\label{sec:dtl}
As discussed above, $\theta_{intra}$ and $\theta_{cross}$ are two crucial hyperparameters that control the precision and recall of the generated pseudo labels. On the one hand, setting these thresholds to small values may lead to the recall of more labels but will inevitably incur some noisy false-positive labels that may hurt the model training process. On the other hand, setting them to large threshold values may cause more positive labels to be missed, resulting in the loss of some data. Thus, it is crucial to exhaustively search for the threshold values that can achieve the optimal performance for different datasets and settings. However, such a process is time-consuming and resource-intensive.

Fortunately, we can also generate pseudo labels for the known labels by Equations \ref{equ:intra-label} and \ref{equ:cross-label} and make full use of the known labels to supervise the threshold learning procedure. To achieve this goal, we propose a DTL algorithm that formulates the nondifferential indication function as a differential formulation to learn the thresholds. For a positive known label $c$ of an image $n$, it is expected that the indicator function is true, and thus, the existence probabilities $\sum_{\{j|y^{n}_{j}=1\}}p_{c,j}^{n} \cdot y^{n}_{j}$ and $\frac{1}{K}\sum_{k=1}^Ks_{c}^{n,k}$ should be larger than the corresponding thresholds. For a negative known label $c$ of image $n$, the two existence probabilities should be smaller than the corresponding thresholds. Thus, we can define the threshold differences by
\begin{gather}
 d^{n, intra}_{c}=(\sum_{\{j|y^{n}_{j}=1\}}p_{c,j}^{n} \cdot y^{n}_{j}) - \theta_{intra}, \\
 d^{n, cross}_{c}=(\frac{1}{K}\sum_{k=1}^Ks_{c}^{n,k}) - \theta_{cross}.
\end{gather}
In this way, we can generate the threshold differences for all known labels and set the threshold differences for the unknown labels to 0, resulting in $\textbf{d}^{n, intra}=\{d^{n, intra}_1, d^{n, intra}_2, \cdots, d^{n, intra}_C\}$ and 

$\textbf{d}^{n, cross}=\{d^{n, cross}_1, d^{n, cross}_2, \cdots, d^{n, cross}_C\}$. Then, we can define the objective losses for optimization as follows:
\begin{gather}
\mathcal{L}_{ist}^{dtl}= \sum_{n=1}^N(\ell^{n}(\textbf{y}^n, \textbf{d}^{n, intra})), \\
\mathcal{L}_{cst}^{dtl}= \sum_{n=1}^N(\ell^{n}(\textbf{y}^n, \textbf{d}^{n, inter})).
\end{gather}
where $\ell^{n}(\textbf{y}^n, \textbf{d}^{n, intra})$ and $\ell^{n}(\textbf{y}^n, \textbf{d}^{n, inter})$ are the partial BCE losses as defined in Equation \ref{eq:pbce}.

\subsection{Optimization}
We follow previous work and use the partial BCE loss as the objective function. Specifically, given the predicted probability distribution $\textbf{p}^n=\{p^n_1, p^n_2, \cdots p^n_C\}$ and the ground-truth known labels, the objective function can be defined as
\begin{equation}
\begin{aligned}
\ell^{n}(\textbf{y}^n, \textbf{p}^n)=&\frac{1}{\sum_{c=1}^C|y^n_c|}\sum_{c=1}^C[\textbf{1}(y^n_c=1)\log(p_c) \\
&+\textbf{1}(y^n_c=-1)\log(1-p_c)].
\label{eq:pbce}
\end{aligned}
\end{equation}
We define a similar objective function for the pseudo labels generated by the IST and CST modules, i.e., $\ell^{n}(\textbf{y}^{n, intra}, \textbf{p}^n)$ and $\ell^{n}(\textbf{y}^{n, cross}, \textbf{p}^n)$, respectively. The final classification loss is defined as the summation of the three losses over all samples, which is formulated as
\begin{equation}
\begin{aligned}
\mathcal{L}_{cls}=&\sum_{n=1}^N(\ell^{n}(\textbf{y}^n, \textbf{p}^n) + \ell^{n}(\hat{\textbf{y}}^{n}, \textbf{p}^n) + \ell^{n}(\tilde{\textbf{y}}^{n}, \textbf{p}^n)).
\end{aligned}
\end{equation}
The final loss can be defined as the sum of the classification loss and the intra-image and cross-image losses:
\begin{equation}
\mathcal{L}=\mathcal{L}_{cls}+\lambda_{1}\mathcal{L}_{ist}+\lambda_{2}\mathcal{L}_{cst} + \lambda_{3}(\mathcal{L}_{ist}^{dtl} + \mathcal{L}_{cst}^{dtl}).
\label{eq:total-loss}
\end{equation}
To ensure all losses contribute to parameter training and may obtain overall well performance, it is expected that different losses share the same amplitude. To this end, $\lambda_{1}$, $\lambda_{2}$ and $\lambda_{3}$ are balance parameters that are set to 10.0, 0.05 and 0.1 in the experiments, respectively.

\section{Experiments}
\label{sec:exp}
In this section, we conduct extensive comparisons with current leading algorithms to demonstrate the superiority of the proposed HST framework. We also perform comprehensive ablation studies to analyze the actual contribution of each component for attaining a better understanding of our approach.

\subsection{Experimental Settings}

\noindent{\textbf{Datasets.}}\quad We follow previous works \cite{durand2019learning} to conduct experiments on the MS-COCO \cite{lin2014microsoft}, Visual Genome \cite{krishna2017visual}, and Pascal VOC 2007 \cite{everingham2010pascal} datasets for evaluation purposes. MS-COCO contains approximately 120k images that cover 80 daily life categories. It is further divided into a training set of approximately 80k images and a validation set of approximately 40k images. Visual Genome contains 108,249 images and covers 80,138 categories. Since most categories have very few samples, we merely consider the 200 most frequent categories, resulting in a VG-200 subset. We randomly select 10,000 images as the test set and the remaining 98,249 images as the training set. Pascal VOC 2007 is the most widely used dataset for multi-label evaluation. It contains approximately 10k images derived from 20 object categories, and these images are divided into a training/validation set of approximately 5,011 images and a test set of 4,952 images.

Because the three datasets are fully annotated, we randomly drop some labels to create a training set with partial labels. In the experiments, the proportion of dropped labels varies from 10\% to 90\%, resulting in 90\% to 10\% known labels.

\noindent{\textbf{Evaluation Metrics.}}\quad For a fair comparison, we follow current works \cite{durand2019learning,huynh2020interactive} and adopt the mean average precision (mAP) over all categories under different proportions of known labels for evaluation purposes. The proportions are set to 10\%, 20\%, $\cdots$, 90\%. We also compute the average mAP over all proportion settings. The overall and per-class F1-measures are also widely used to evaluate MLR models \cite{chen2019learning}, and we additionally follow previous works \cite{chen2018recurrent} to adopt the OF1 and CF1 metrics for a more comprehensive evaluation.

\noindent{\textbf{Implementation Details.}}\quad To fairly compare our approach with existing algorithms, we follow previous works \cite{durand2019learning,chen2019learning} and adopt the 101-layer ResNet \cite{he2016deep} as the backbone to extract global feature maps $\textbf{f}^n$. Then, we utilize $\phi_{sarl}(\cdot)$ to learn $[\textbf{f}^n_1, \textbf{f}^n_2, \cdots, \textbf{f}^n_C]$. The classifier $\phi(\cdot)$ is implemented with a gated graph neural network to explore contextual information, and this is followed by a fully connected layer to produce the score vector. The co-occurrence estimation function $\phi_{intra}(\cdot)$ is implemented by three fully connected layers, among which the first layer maps the 1024-dimension vector to a 512-dimension vector and applies a rectified linear unit (ReLU) function, the second layer maps the 512-dimension vector to a 1,024-dimension vector and is also followed by a ReLU function, and the last layer maps to a score that indicates the co-occurrence probability.

\begin{table*}
  \centering
  \small
  \begin{tabular}{c|c|ccccccccc|c}
  \hline
  \centering Datasets & Methods & 10\% & 20\% & 30\% & 40\% & 50\% & 60\% & 70\% & 80\% & 90\% & Ave. mAP \\
  \hline
  \hline
  \centering \multirow{8}*{MS-COCO} & SSGRL & 62.5 & 70.5 & 73.2 & 74.5 & 76.3 & 76.5 & 77.1 & 77.9 & 78.4 & 74.1 \\
  \centering ~ & GCN-ML & 63.8 & 70.9 & 72.8 & 74.0 & 76.7 & 77.1 & 77.3 & 78.3 & 78.6 & 74.4 \\
  \centering ~ & KGGR & 66.6 & 71.4 & 73.8 & 76.7 & 77.5 & 77.9 & 78.4 & 78.7 & 79.1 & 75.6 \\
  \centering ~ & P-GCN & 67.5 & 71.6 & 73.8 & 75.5 & 77.4 & 78.4 & 79.5 & \textbf{80.7} & \textbf{81.5} & 76.2 \\
  \centering ~ & ASL & 69.7 & 74.0 & 75.1 & 76.8 & 77.5 & 78.1 & 78.7 & 79.1 & 79.7 & 76.5 \\
  \centering ~ & CL & 26.7 & 31.8 & 51.5 & 65.4 & 70.0 & 71.9 & 74.0 & 77.4 & 78.0 & 60.7 \\
  \centering ~ & Partial BCE & 61.6 & 70.5 & 74.1 & 76.3 & 77.2 & 77.7 & 78.2 & 78.4 & 78.5 & 74.7 \\
  \centering ~ & SST & 68.1 & 73.5 & 75.9 & 77.3 & 78.1 & 78.9 & 79.2 & 79.6 & 79.9 & 76.7 \\
  \centering ~ & HST & \textbf{70.6} & \textbf{75.8} & \textbf{77.3} & \textbf{78.3} & \textbf{79.0} & \textbf{79.4} & \textbf{79.9} & 80.2 & 80.4 & \textbf{77.9} \\
  \hline
  \hline
  \centering \multirow{8}*{VG-200} & SSGRL & 34.6 & 37.3 & 39.2 & 40.1 & 40.4 & 41.0 & 41.3 & 41.6 & 42.1 & 39.7 \\
  \centering ~ & GCN-ML & 32.0 & 37.8 & 38.8 & 39.1 & 39.6 & 40.0 & 41.9 & 42.3 & 42.5 & 39.3 \\
  \centering ~ & KGGR & 36.0 & 40.0 & 41.2 & 41.5 & 42.0 & 42.5 & 43.3 & 43.6 & 43.8 & 41.5 \\
  \centering ~ & P-GCN & - & - & - & - & - & - & - & - & - & - \\
  \centering ~ & ASL & 38.4 & 39.5 & 40.6 & 40.6 & 41.1 & 41.4 & 41.4 & 41.6 & 41.6 & 40.7 \\
  \centering ~ & CL & 12.1 & 19.1 & 25.1 & 26.7 & 30.0 & 31.7 & 35.3 & 36.8 & 38.5 & 28.4 \\
  \centering ~ & Partial BCE & 27.4 & 38.1 & 40.2 & 40.9 & 41.5 & 42.1 & 42.4 & 42.7 & 42.7 & 39.8 \\
  \centering ~ & SST & 38.8 & 39.4 & 41.1 & 41.8 & 42.7 & 42.9 & 43.0 & 43.2 & 43.5 & 41.8 \\
  \centering ~ & HST & \textbf{40.6} & \textbf{41.6} & \textbf{43.3} & \textbf{44.6} & \textbf{45.2} & \textbf{45.8} & \textbf{46.8} & \textbf{47.2} & \textbf{47.8} & \textbf{44.8} \\
  \hline
  \hline
  \centering \multirow{8}*{VOC 2007} & SSGRL & 77.7 & 87.6 & 89.9 & 90.7 & 91.4 & 91.8 & 92.0 & 92.2 & 92.2 & 89.5 \\
  \centering ~ & GCN-ML & 74.5 & 87.4 & 89.7 & 90.7 & 91.0 & 91.3 & 91.5 & 91.8 & 92.0 & 88.9 \\
  \centering ~ & KGGR & 81.3 & 88.1 & 89.9 & 90.4 & 91.2 & 91.3 & 91.5 & 91.6 & 91.8 & 89.7 \\
  \centering ~ & P-GCN & 82.5 & 85.4 & 88.2 & 89.8 & 90.0 & 90.9 & 91.6 & 92.5 & \textbf{93.1} & 89.3 \\
  \centering ~ & ASL & 82.9 & 88.6 & 90.0 & 91.2 & 91.7 & 92.2 & 92.4 & 92.5 & 92.6 & 90.5 \\
  \centering ~ & CL & 44.7 & 76.8 & 88.6 & 90.2 & 90.7 & 91.1 & 91.6 & 91.7 & 91.9 & 84.1 \\
  \centering ~ & Partial BCE & 80.7 & 88.4 & 89.9 & 90.7 & 91.2 & 91.8 & 92.3 & 92.4 & 92.5 & 90.0 \\
  \centering ~ & SST & 81.5 & 89.0 & 90.3 & 91.0 & 91.6 & 92.0 & 92.5 & 92.6 & 92.7 & 90.4 \\
  \centering ~ & HST & \textbf{84.3} & \textbf{89.1} & \textbf{90.5} & \textbf{91.6} & \textbf{92.1} & \textbf{92.4} & \textbf{92.5} & \textbf{92.8} & 92.8 & \textbf{90.9} \\
  \hline
  \end{tabular}
  \vspace{10pt}
  \caption{The average mAP and mAP values achieved by our HST framework and current state-of-the-art competitors with different known label proportions for MLR-PL on the MS-COCO, VG-200 and Pascal VOC 2007 datasets. An ``-" denotes the corresponding result is not provided. The best results are highlighted in bold.}
  \vspace{0pt}
  \label{tab:mAP-results}
\end{table*}

\begin{table}[!t]
  \centering
  \footnotesize
  \begin{tabular}{c|c|cc}
  \hline
  \centering Datasets & Methods & Avg. OF1  & Avg. CF1 \\
  \hline
  \hline
  \centering \multirow{8}*{MS-COCO} & SSGRL & 73.9 & 68.1 \\
  \centering ~ & GCN-ML & 73.1 & 68.4 \\
  \centering ~ & KGGR & 73.7 & 69.7 \\
  \centering ~ & ASL & 46.7 &  47.9 \\
  \centering ~ & CL & 61.9 & 48.3 \\
  \centering ~ & Partial BCE & 74.0 & 68.8 \\
  \centering ~ & SST & 75.8 & 71.2 \\
  \centering ~ & HST & \textbf{76.7} & \textbf{72.6} \\
  \hline
  \hline
  \centering \multirow{8}*{VG-200} & SSGRL & 37.8 & 26.1 \\
  \centering ~ & GCN-ML & 38.7 & 25.6 \\
  \centering ~ & KGGR & 41.2 & 33.6 \\
  \centering ~ & ASL & 24.9 &  23.3 \\
  \centering ~ & CL & 23.6 & 10.9 \\
  \centering ~ & Partial BCE & 36.1 &  25.7 \\
  \centering ~ & SST & 39.9 & 30.8 \\
  \centering ~ & HST & \textbf{46.3} & \textbf{37.9} \\
  \hline
  \hline
  \centering \multirow{8}*{VOC 2007} & SSGRL & 87.7 & 84.5 \\
  \centering ~ & GCN-ML & 87.3 & 84.6 \\
  \centering ~ & KGGR & 86.5 & 84.7 \\
  \centering ~ & ASL & 41.0 & 40.9 \\
  \centering ~ & CL & 83.8 &75.4 \\
  \centering ~ & Partial BCE & 87.9 & 84.8 \\
  \centering ~ & SST & 88.2 & 85.6 \\
  \centering ~ & HST & \textbf{88.4} & \textbf{86.1} \\
  \hline
  \end{tabular}
  \vspace{2pt}
  \caption{The average OF1 and CF1 values of our HST framework and current state-of-the-art competitors for MLR-PL on the MS-COCO, VG-200, and Pascal VOC 2007 datasets. The best results are highlighted in bold.}
  \label{tab:average-results}
\end{table}

The proposed framework is trained by using the loss $\mathcal{L}$, as shown in Equation \ref{eq:total-loss}. The parameters of ResNet-101 are initialized by those pre-trained on the ImageNet \cite{deng2009imagenet} dataset, and the parameters of all other layers are initialized randomly. The model is trained via the adaptive moment estimation (ADAM) algorithm \cite{kingma2015adam} with a batch size of 32, momentums of 0.999 and 0.9, and a weight decay of $5 \times 10^{-4}$. The original learning rate is set to 0.00001, and it is divided by 10 every 10 epochs. The model is trained with 20 epochs in total. During training, the input image is resized to 512$\times$512, and we randomly choose a number from \{512, 448, 384, 320, 256\} as the width and height to crop the image to a patch. Finally, the cropped patch is further resized to 448$\times$448. Then, we perform random horizontal flipping and normalization. $\theta_{intra}$ and $\theta_{cross}$ are two crucial parameters that control the accuracy of the generated pseudo labels. During the training process, the parameters are set to 1 for the first 5 epochs to avoid incurring any pseudo labels.

During inference, the IST and CST modules are removed, and the image is resized to 448$\times$448 for evaluation purposes.

\subsection{Comparison with State-of-the-Art Methods}
To evaluate the effectiveness of the proposed HST framework, we compare it with the following algorithms, which can be classified into three groups. 1) \textbf{SSGRL} (ICCV'19) \cite{chen2019learning}, knowledge-guided graph routing (\textbf{KGGR}; TPAMI'22) \cite{chen2022knowledge}, graph convolution with machine learning (\textbf{GCN-ML}; CVPR'19) \cite{chen2019multi} and \textbf{P-GCN} (TPAMI'21) \cite{chen2021p-gcn} introduce graph neural networks to model label dependencies, and they achieve state-of-the-art performance on the traditional MLR task. We adapt these methods to address MLR-PL by replacing the loss with the partial BCE loss while keeping other components unchanged. 2) \textbf{CL} (CVPR'19) \cite{durand2019learning} alternately labels the unknown labels with strong evidence to update the training set and retrains the model with an updated training set. We also treat this approach as a strong baseline for addressing the MLR-PL task. 3) \textbf{Partial BCE} (CVPR'19) \cite{durand2019learning} and \textbf{ASL} (ICCV'21) \cite{ridnik2021asymmetric} were proposed to address the MLR-PL and MLR tasks, respectively, by designing new loss functions. The former introduces a normalized BCE loss to better exploit partial labels for training the multi-label models, and the latter introduces a novel asymmetric loss based on the focal loss \cite{lin2017focal-loss} to better balance negative and positive samples in the traditional MLR task. For fair comparisons, we adopt the same ResNet-101 network as the backbone and follow the exact same train/validation split settings.

\subsubsection{Performance on MS-COCO}
The mAP, OF1, and CF1 measures are the most important metrics for evaluating MLR performance. As shown in Tables \ref{tab:mAP-results} and \ref{tab:average-results}, our HST framework obtains average mAP, OF1, CF1 values of 77.9\%, 76.7\%, and 72.6\% on the MS-COCO dataset, with average improvements of 1.2\%, 0.9\%, and 1.4\%, respectively. In addition to these average results, we further present the mAPs achieved on all proportion settings, as shown in Table \ref{tab:mAP-results}. We find that the traditional MLR algorithms can achieve competitive performance when the proportion of known labels is high (e.g., 80\%-90\%), but they suffer from obvious performance drops when the proportion decreases (e.g., 10\%-20\%). ASL adaptively balances the loss weights for the negative and positive samples and achieves very good performance across different known label settings. By exploiting the intra-image and cross-image semantic correlations to generate pseudo labels for unknown labels, our HST framework obtains the best performance for almost all known label proportion settings. Specifically, it obtains mAPs of 70.6\%, 75.8\%, 77.3\%, 78.3\%, 79.0\%, 79.4\%, 79.9\%, 80.2\%, and 80.4\% under the settings of 10\%-90\% known labels, outperforming the second-best ASL algorithm by 0.9\%, 1.8\%, 2.2\%, 1.5\%, 1.5\%, 1.3\%, 1.2\%, 1.1\%, and 0.7\%, respectively. Notably, SST is the preliminary version of this research. The extension of SST to HST demonstrate improved performance across all evaluated metrics.

\subsubsection{Performance on VG-200}
Compared to MS-COCO, VG-200 is a more challenging dataset that covers more categories. Here, we also present the comparisons performed on this dataset to provide more comprehensive evaluations. As shown in Tables \ref{tab:mAP-results} and \ref{tab:average-results}, our HST framework obtains the best average mAP, OF1, and CF1 values of 44.8\%, 46.3\%, and 37.9\%, respectively, so it demonstrates a more obvious improvement over the current leading algorithms. As shown in Table \ref{tab:mAP-results}, we find that some algorithms (e.g., KGGR and GCN-ML) obtain better results under the high known label proportion settings, while other algorithms (e.g., ASL) perform better under the low known label proportion settings. The HST framework still obtains the best performance for all settings. Concretely, the HST framework obtains 4.6\% and 8.6\% mAP improvements over the KGGR and GCN-ML algorithms under the 10\% known label proportion setting and obtains a 6.2\% mAP improvement over ASL under the 90\% known label proportion setting. Similarly, HST exhibits a significant enhancement over SST, achieving improvements in average mAP, OF1, and CF1 scores by 3.0\%, 6.4\%, and 7.1\%, respectively.

\subsubsection{Performance on Pascal VOC 2007}
Pascal VOC 2007 is the most widely used dataset for MLR, and we also present the results obtained on this dataset in Tables \ref{tab:mAP-results} and \ref{tab:average-results}. As this dataset covers only 20 categories and is simpler than Visual Genome and MS-COCO, current algorithms achieve comparable results when utilizing certain proportions of known labels (e.g., more than 40\%). However, the performance also drops dramatically when the proportion decreases to 10\% and 20\%. In contrast, our proposed HST framework consistently outperforms the current methods under all proportion settings. Specifically, it outperforms the traditional multi-label methods (i.e., KGGR and P-GCN) and the second-best SST method by 3.0\%, 1.8\%, and 2.8\% in term of mAP when the known label proportion is 10\%.

\subsection{Ablation Studies}
In this section, we present comprehensive experiments to analyze the actual contribution of each crucial component. Here, we mainly present the average mAP, OF1, and CF1 as they can better describe the overall performance. 

\begin{table}[!t]
  \centering
  \scriptsize
  \begin{tabular}{c|c|ccc}
  \hline
  \centering \multirow{2}*{Datasets} & \multirow{2}*{Methods} & \multicolumn{3}{c}{Average Results} \\
  \cline{3-5}
  \centering ~ & ~ & mAP & OF1 & CF1 \\
  \hline
  \hline
  \centering \multirow{3}*{MS-COCO}  & SSGRL & 74.1 & 73.9 & 68.1  \\
  \centering ~ & SSGRL w/ PL & 75.3 & 74.8 & 69.5 \\
  \centering ~ & Ours & \textbf{77.9} & \textbf{76.7} & \textbf{72.6} \\
  \hline
  \hline
  \centering \multirow{3}*{VG-200} & SSGRL & 39.7 & 37.8 & 26.1 \\
  \centering ~ & SSGRL w/ PL & 40.8 & 41.4 & 33.1 \\
  \centering ~ & Ours & \textbf{44.8} & \textbf{46.3} & 37.9 \\
  \hline
  \hline
  \centering \multirow{3}*{VOC 2007} & SSGRL & 89.5 & 87.7 & 84.5 \\
  \centering ~ & SSGRL w/ PL & 89.6 & 87.4 & 85.1 \\
  \centering ~ & Ours & \textbf{90.9} & \textbf{88.4} & \textbf{86.1} \\
  \hline
  \end{tabular}
  \vspace{2pt}
  \caption{The average OF1 and CF1 values of the baseline SSGRL method, SSGRL with confidence-based pseudo label generation (SSGRL w/ PL) and our framework (Ours) on the MS-COCO, VG-200 and Pascal VOC 2007 datasets. The best results are highlighted in bold.}
  \label{tab:average-ablation-results-hst}
\end{table}

\subsubsection{Analysis of HST}
As mentioned above, HST uses the SD algorithm to implement the SARL module, and thus, SSGRL can be regarded as the baseline. Here, we emphasize the comparisons with the SSGRL method to verify the effectiveness of the whole HST module. As shown in Table \ref{tab:average-ablation-results-hst}, the baseline SSGRL method obtains average mAP values of 74.1\%, 39.7\%, 89.5\% on the MS-COCO, VG-200, and Pascal VOC 2007 datasets, respectively. By integrating the HST module to complement the unknown labels, the average mAP values are boosted to 77.9\%, 44.8\%, 90.9\%, with improvements of 3.8\%, 5.1\%, and 1.4\%, respectively. Similar improvements can be observed for other metrics, such as the average OF1 and CF1 values, as shown in Table \ref{tab:average-ablation-results-hst}.

HST introduces intra- and cross-image semantic correlations to help generate pseudo labels for the unknown labels. To further verify the effectiveness of these semantic correlations, we implement another baseline that simply generates pseudo labels based on the predicted confidence scores (denoted as ``SSGRL w/ PL") for comparison purposes. As shown in Tables \ref{tab:average-ablation-results-hst}, ``SSGRL w/ PL" obtains slightly better performance than SSGRL. However, the proposed HST framework performs better than this baseline, with average mAP improvements of 2.6\%, 4.0\%, and 1.3\% on the MS-COCO, VG-200, and Pascal VOC 2007 datasets, respectively. Similarly, ``SSGRL w/ PL" also achieves slight performance improvements but is still worse than the proposed HST framework in terms of other metrics, such as OF1 and CF1 in Table \ref{tab:average-ablation-results-hst}.

Since the HST framework consists of two complementary modules, i.e., the IST module and the CST module, in the following, we further conduct more ablation experiments to analyze the actual contribution of each module for a more in-depth understanding.

\begin{table}[!t]
  \centering
  \scriptsize
  \begin{tabular}{c|c|ccc}
  \hline
  \centering \multirow{2}*{Datasets} & \multirow{2}*{Methods} & \multicolumn{3}{c}{Average Results} \\
  \cline{3-5}
  \centering ~ & ~ & mAP & OF1 & CF1 \\
  \hline
  \hline
  \centering \multirow{5}*{MS-COCO} & SSGRL & 74.1 & 73.9 & 68.1  \\
  \centering ~ & Ours IST & 76.4 & 75.5 & 70.7 \\
  \centering ~ & Ours IST w/ stat & 70.7 & 71.0 & 63.4 \\
  \centering ~ & Ours IST w/o $\mathcal{L}_{ist}$ & 75.5 & 75.2 & 70.4 \\
  \centering ~ & Ours & \textbf{77.9} & \textbf{76.7} & \textbf{72.6} \\
  \hline
  \hline
  \centering \multirow{5}*{VG-200} & SSGRL & 39.7 & 37.8 & 26.1 \\
  \centering ~ & Ours IST & 42.4 & 42.2 & 32.8 \\
  \centering ~ & Ours IST w/ stat & 39.6 & 37.4 & 27.8 \\
  \centering ~ & Ours IST w/o $\mathcal{L}_{ist}$ & 41.6 & 42.6 & 32.3 \\
  \centering ~ & Ours & \textbf{44.8} & \textbf{46.3} & \textbf{37.9} \\
  \hline
  \hline
  \centering \multirow{5}*{VOC 2007} & SSGRL & 89.5 & 87.7 & 84.5 \\
  \centering ~ & Ours IST & 90.0 & 87.3 & 84.3 \\
  \centering ~ & Ours IST w/ stat & 86.7 & 83.3 & 81.0 \\
  \centering ~ & Ours IST w/o $\mathcal{L}_{ist}$ & 89.0 & 87.0 & 83.9 \\
  \centering ~ & Ours & \textbf{90.9} & \textbf{88.4} & \textbf{86.1} \\
  \hline
  \end{tabular}
  \vspace{2pt}
  \caption{The average mAP, OF1, and CF1 values of the baseline SSGRL method, our framework with only the IST module (Ours IST), our framework utilizing only the IST model with statistical co-occurrence (Ours IST w/ stat), our framework employing only the IST module without the loss $\mathcal{L}_{ist}$ (Ours IST w/o $\mathcal{L}_{ist}$) our framework (Ours) on the MS-COCO, VG-200 and Pascal VOC 2007 datasets. The best results are highlighted in bold.}
  \label{tab:average-ablation-results-ist}
\end{table}

\subsubsection{Analysis of the IST Module}
To evaluate the actual contribution of the IST module, we compare the performance of our HST framework utilizing only this module with the performance of the baseline SSGRL method. As presented in Table \ref{tab:average-ablation-results-ist}, ``Ours IST" achieves obvious performance improvements over the baseline SSGRL method, i.e., average mAP improvements of 2.3\%, 2.7\%, and 0.5\% on the MS-COCO, VG-200 and Pascal VOC 2007 datasets, respectively. In addition, as shown in Table \ref{tab:average-ablation-results-ist}, ``Ours IST" also outperforms the baseline SSGRL method in terms of the OF1 and CF1 metrics; i.e., it achieves average OF1 improvements of 1.6\% and 4.4\% and average CF1 improvements of 2.6\%, 6.7\% on the MS-COCO and VG-200 datasets, respectively. These results demonstrate that the semantic correlations within images can be helpful for complementing unknown labels.

In the IST module, the loss $\mathcal{L}_{ist}$ helps to learn an accurate image-specific co-occurrence matrix for generating highly confident pseudo labels. To evaluate the contribution of this module, we conduct experiments that only utilize the IST module without the loss for comparison purposes (namely, ``Ours IST w/o $\mathcal{L}_{ist}$"). As shown in Table \ref{tab:average-ablation-results-ist}, it decreases the average mAPs by 0.9\%, 0.8\%, and 1.0\% on the MS-COCO, VG-200, and Pascal VOC 2007 datasets, respectively.

To better model label co-occurrences, we propose to learn an image-specific co-occurrence matrix. To verify the contribution of this matrix, we conduct an experiment that uses the statistical co-occurrences computed on the training set to replace the learned image-specific co-occurrence matrix, with the other aspects unchanged (namely, ``Ours IST w/ stat"). As shown in Tables \ref{tab:average-ablation-results-ist}, this version suffers from dramatic performance drops in terms of all metrics, especially in scenarios with low known label proportions. For example, the mAP is merely 58.6\%, an extremely obvious degradation of 8.6\% compared with that of ``Ours IST" on the MS-COCO dataset with 10\% known labels. One reason for this phenomenon is that statistical co-occurrences are not suitable for every image, and thus, this approach may incur many false-positive labels for unsuitable images.

\begin{figure}[!t]
   \centering
   \includegraphics[width=0.98\linewidth]{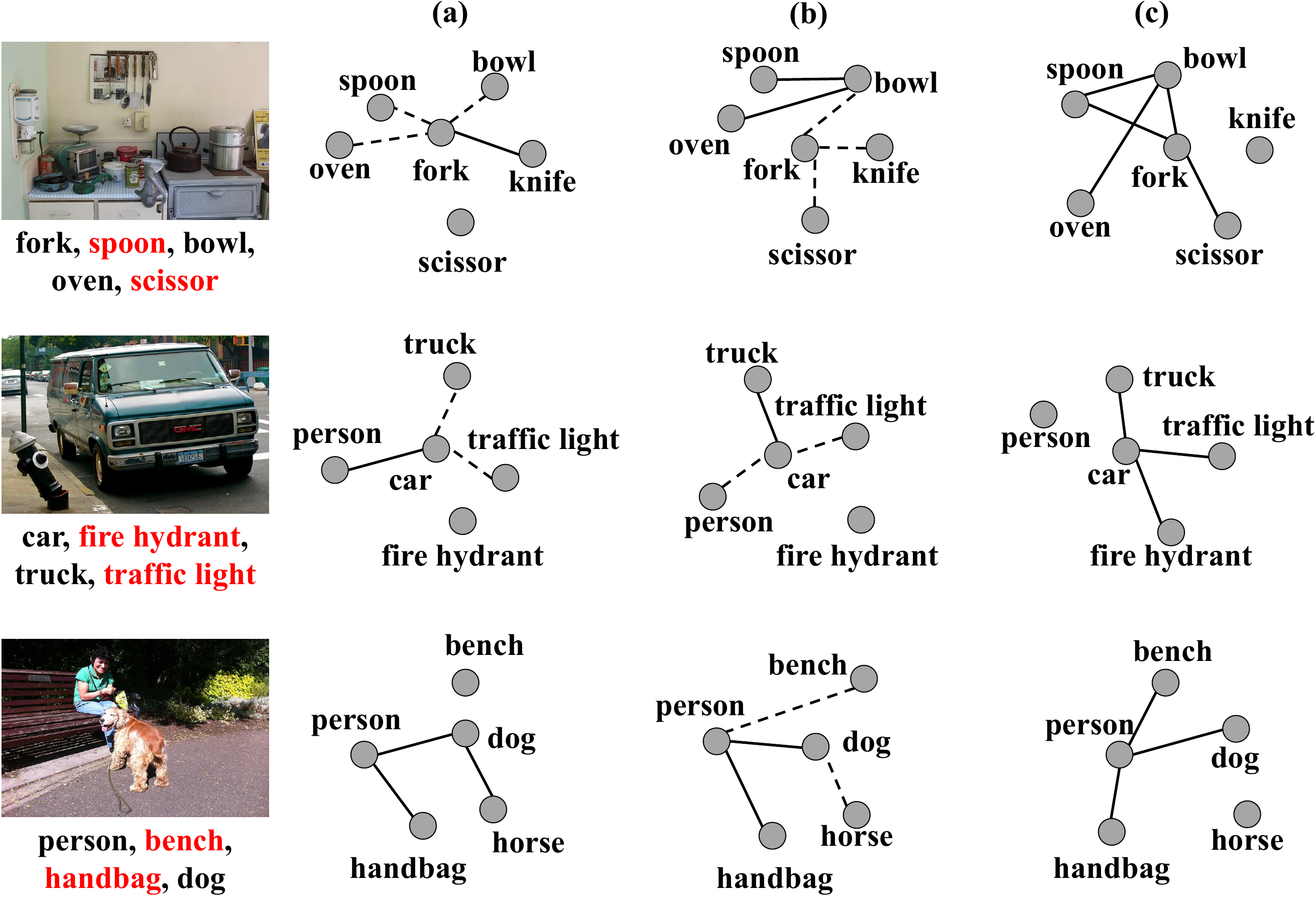}
   \caption{Several examples of images with partial labels (unknown labels are highlighted in red) and their corresponding co-occurrence matrices: (a) the statistical co-occurrence matrix, (b) the image-specific co-occurrence matrix generated by IST without $\mathcal{L}_{ist}$, (c) the image-specific co-occurrence matrix generated by IST.}
   \label{fig:vis-ist}
\end{figure}

To delve deep into the IST module, we further visualize some examples with co-occurrence matrices in Figure \ref{fig:vis-ist}. As shown, a statistical co-occurrence matrix can provide some common co-occurring pair information to complement unknown labels (e.g., ``person" and ``handbag"; ``fork" and ``knife"), but it is not suitable for each image. For instance, given the first example, the statistical co-occurrence matrix will output a high confidence value for "knife" based on the existence of "fork", which results in false-positive labels. Similarly, the statistical co-occurrence matrix will miss many potential co-occurrence pairs because of the imbalance induced when collecting training samples. By learning image-specific co-occurrences, the IST module captures the category pairs that frequently co-occur, such as ``car" and ``fire hydrant" in the second example in Figure \ref{fig:vis-ist}. This demonstrates that learning image-specific co-occurrences can better capture label correlations for each image and thus facilitate the generation of more accurate pseudo labels.

\begin{table}[!t]
  \centering
  \scriptsize
  \begin{tabular}{c|c|ccc}
  \hline
  \centering \multirow{2}*{Datasets} & \multirow{2}*{Methods} & \multicolumn{3}{c}{Average Results} \\
  \cline{3-5}
  \centering ~ & ~ & mAP & OF1 & CF1 \\
  \hline
  \hline
  \centering \multirow{5}*{MS-COCO}  & SSGRL & 74.1 & 73.9 & 68.1  \\
  \centering ~ & Ours CST &  76.5 & 74.6 & 69.8 \\
  \centering ~ & Ours CST w/ IL & 75.5 & 70.9 & 69.2 \\
  \centering ~ & Ours CST w/o $\mathcal{L}_{cst}$ & 74.8 & 72.7 & 69.9 \\
  \centering ~ & Ours & \textbf{77.9} & \textbf{76.7} & \textbf{72.6} \\
  \hline
  \hline
  \centering \multirow{5}*{VG-200} & SSGRL & 39.7 & 37.8 & 26.1 \\
  \centering ~ & Ours CST & 42.7 & 42.5 & 35.4 \\
  \centering ~ & Ours CST w/ IL & 42.1 & 41.7 & 34.0 \\
  \centering ~ & Ours CST w/o $\mathcal{L}_{cst}$ & 42.1 & 41.6 & 33.5 \\
  \centering ~ & Ours & \textbf{44.8} & \textbf{46.3} & 37.9 \\
  \hline
  \hline
  \centering \multirow{5}*{VOC 2007} & SSGRL & 89.5 & 87.7 & 84.5 \\
  \centering ~ & Ours CST & 90.4 & 87.9 & 85.4 \\
  \centering ~ & Ours CST w/ IL & 90.0 & 87.6 & 85.0 \\
  \centering ~ & Ours CST w/o $\mathcal{L}_{cst}$ & 89.9 & 87.5 & 84.9 \\
  \centering ~ & Ours & \textbf{90.9} & \textbf{88.4} & \textbf{86.1} \\
  \hline
  \end{tabular}
  \vspace{2pt}
  \caption{The average mAP, OF1, and CF1 values of the baseline SSGRL method, our framework with only the CST module (Ours CST), our framework utilizing only the CST module with instance-level features (Ours CST w/ IL), our framework employing only the CST module without the loss $\mathcal{L}_{cst}$ (Ours CST w/o $\mathcal{L}_{cst}$) and our framework (Ours) on the MS-COCO, VG-200 and Pascal VOC 2007 datasets. The best results are highlighted in bold.}
  \label{tab:average-ablation-results-cst}
\end{table}

\subsubsection{Analysis of the CST Module}
\label{sec:exp-cst}
To evaluate the actual contribution of the CST module, we add the CST module to the baseline SSGRL method, namely, ``Ours CST", and compare it with the baseline SSGRL approach. As shown in Table \ref{tab:average-ablation-results-cst}, the CST module improves the average mAPs from 74.1\%, 39.7\%, and 89.5\% to 76.5\%, 42.7\%, and 90.4\% on the MS-COCO, VG-200, Pascal VOC 2007 datasets, with average mAP improvements of 2.4\%, 3.0\%, and 0.9\%, respectively. Similar to the IST module, ``Ours CST" consistently outperforms the baseline SSGRL method in terms of the OF1 and CF1 metrics; i.e., it provides average OF1 improvements of 0.7\%, 4.7\%, and 0.2\% and average CF1 improvements of 1.7\%, 9.3\%, and 0.9\% on the MS-COCO, VG-200 and Pascal VOC 2007 datasets, respectively. 

\begin{figure}[!t]
   \centering
   \includegraphics[width=0.95\linewidth]{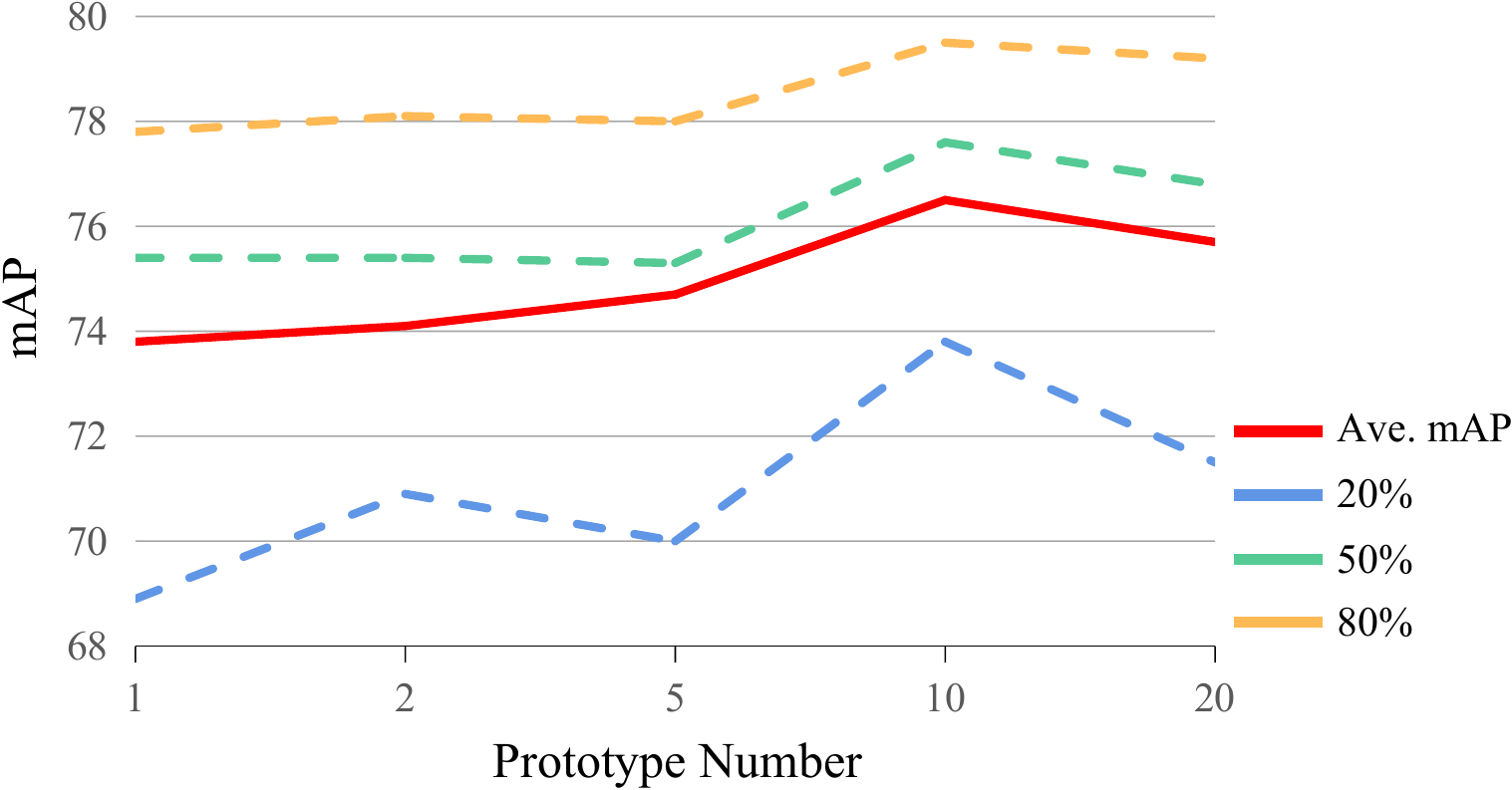}
   \caption{The performance of the HST framework with only the CST module under different numbers of prototypes. We present the average mAP over all known label proportion settings and the mAPs achieved under 20\%, 50\%, 80\% known label settings on the MS-COCO dataset.}
   \label{fig:prototypenum}
\end{figure}

As with the IST module, the loss $\mathcal{L}_{cst}$ helps the CST module to obtain compact representation prototypes. Here, we conduct experiments that merely utilize the CST module without this loss for comparison purposes (namely, ``Ours CST w/o $\mathcal{L}_{cst}$") to verify its effect. As shown in Table \ref{tab:average-ablation-results-cst}, it decreases the average mAPs by 1.0\%, 0.6\%, and 0.4\% on the MS-COCO, VG-200, and Pascal VOC 2007 datasets, respectively.

\begin{figure*}[!t]
\centering
\subfigure{
\includegraphics[width=0.48\linewidth]{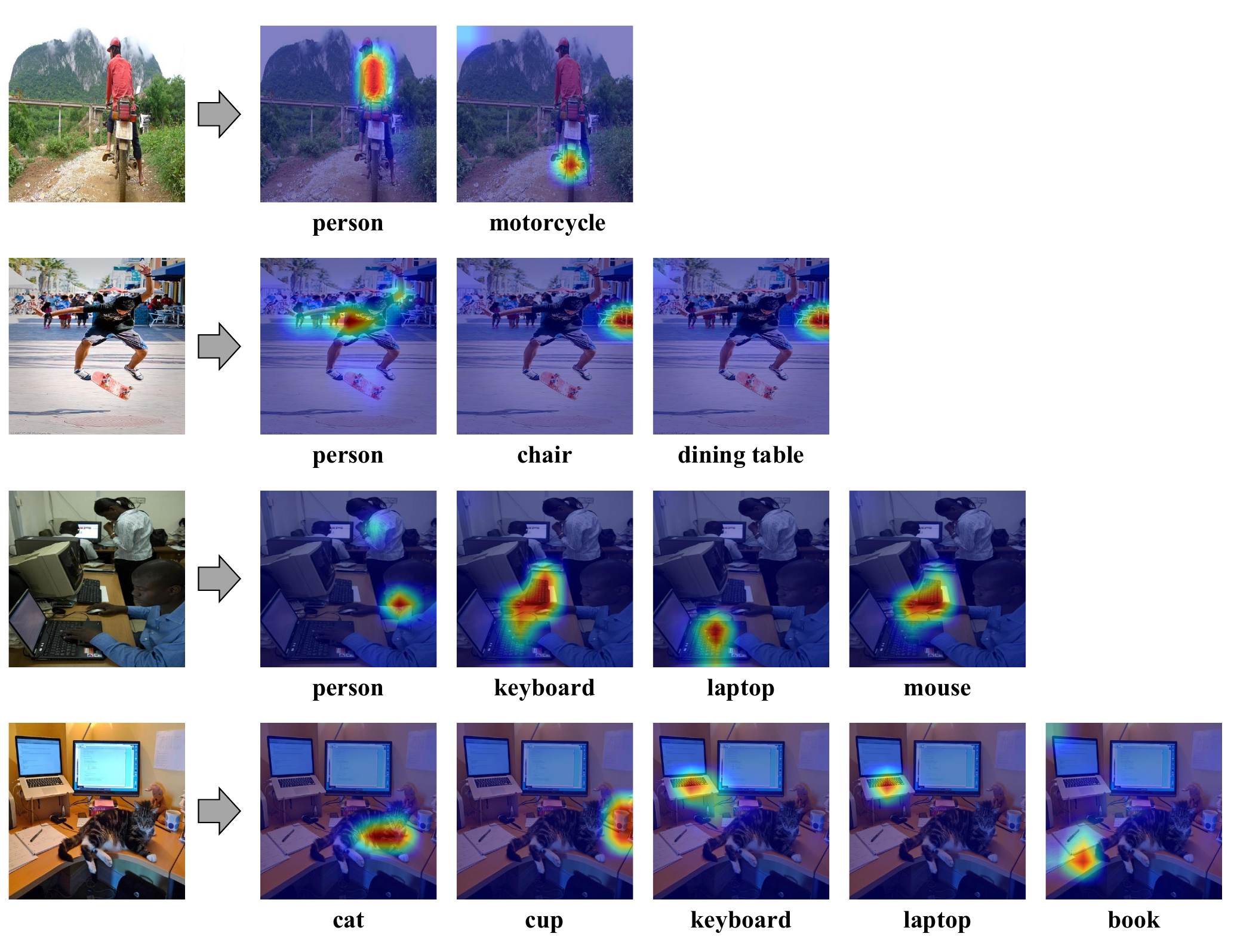}}
\subfigure{
\includegraphics[width=0.48\linewidth]{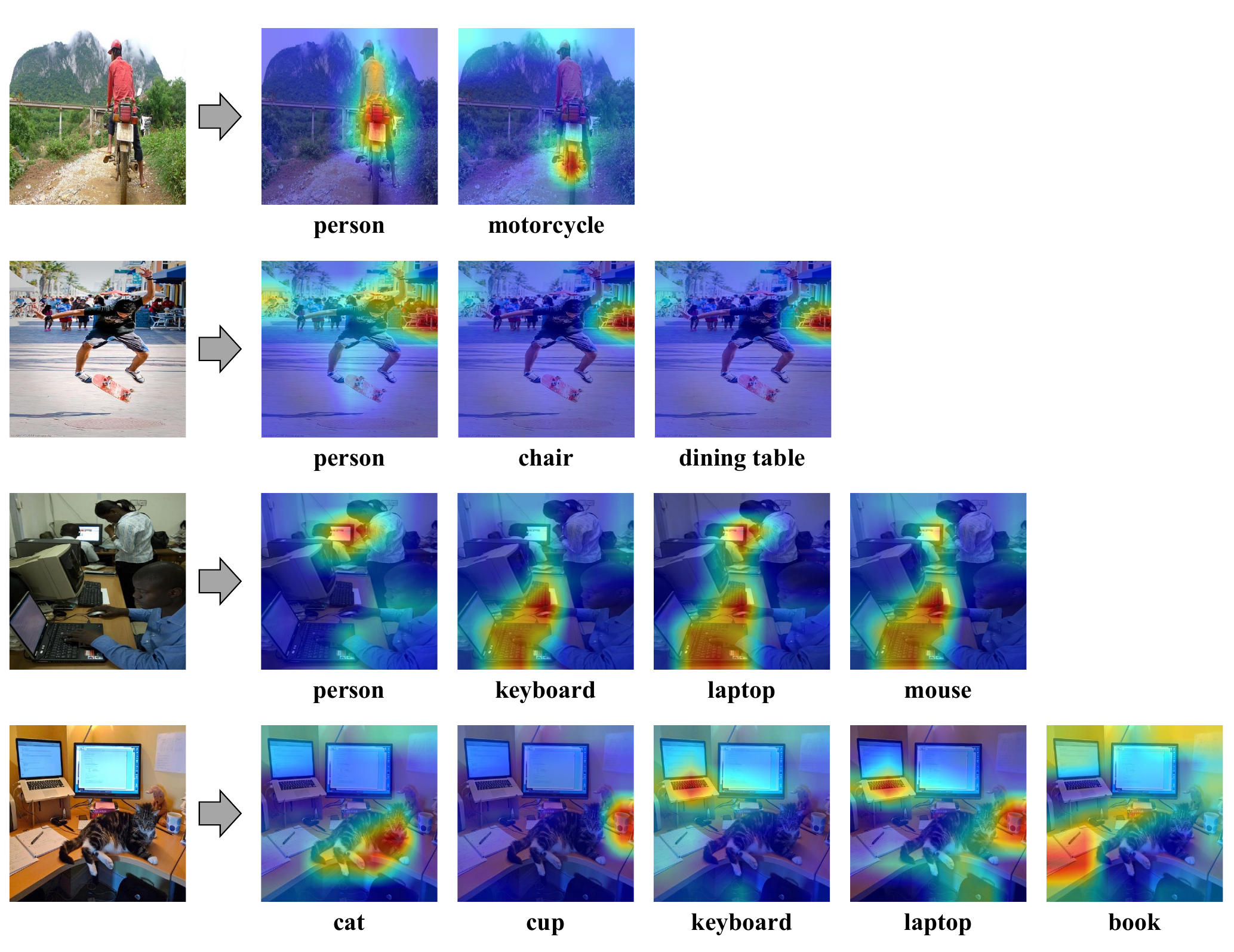}}
\vspace{-10pt}
\caption{Some examples of the input images and semantic maps corresponding to the categories existing in the images generated by SD (Left) and SA (right), respectively.}
\label{fig:visualization}
\end{figure*}

In the conference version, we utilized instance-level feature similarities to help generate pseudo labels. Here, we also conduct experiments to compare the performance of the proposed instance-prototype and instance-level similarities. As presented in Tables \ref{tab:average-ablation-results-cst}, replacing the instance-prototype similarities with instance-level similarities decreases the values of all metrics. For example, this version suffers from average mAPs drop of 1.0\%, 0.6\%, and 0.4\% on the MS-COCO, VG-200, and Pascal VOC 2007 datasets, respectively. One possible reason for these phenomena is that prototype representation can better describe the overall information of each category and thus can better measure the correlations between each instance and the corresponding categories.

As discussed above, we learn $K$ prototypes for each category, and here, we conduct experiments that vary the number of prototypes $K$ from 1 to 20 to find the optimal number. As shown in Figure \ref{fig:prototypenum}, increasing $K$ from 1 to 10 can improve the average mAP and the mAPs under different known label proportions, but increasing $K$ to 20 decreases the performance. Thus, we set $K$ as 10 in all the experiments.

\begin{table}[!t]
  \centering
  \scriptsize
  \begin{tabular}{c|c|ccc}
  \hline
  \centering \multirow{2}*{Datasets} & \multirow{2}*{Methods} & \multicolumn{3}{c}{Average Results} \\
  \cline{3-5}
  \centering ~ & ~ & mAP & OF1 & CF1 \\
  \hline
  \hline
  \centering \multirow{2}*{MS-COCO} & Ours w/ SA & 77.7 & 76.2 & 72.4 \\
  \centering ~ & Ours & \textbf{77.9} & \textbf{76.7} & \textbf{72.6} \\
  \hline
  \hline
  \centering \multirow{2}*{VG-200} & Ours w/ SA & 44.4 & 44.8 & \textbf{38.5} \\
  \centering ~ & Ours & \textbf{44.8} & \textbf{46.3} & 37.9 \\
  \hline
  \hline
  \centering \multirow{2}*{VOC 2007} & Ours w/ SA & 90.7 & 87.5 & 85.0 \\
  \centering ~ & Ours & \textbf{90.9} & \textbf{88.4} & \textbf{86.1} \\
  \hline
  \end{tabular}
  \vspace{2pt}
  \caption{The average OF1 and CF1 values of our framework with SA instead of SD (Ours w/ SA) and our framework (Ours) on the MS-COCO, VG-200 and Pascal VOC 2007 datasets. The best results are highlighted in bold.}
  \label{tab:average-ablation-results-sarl}
\end{table}

\begin{table*}[!t]
  \centering
  \small
  \begin{tabular}{c|c|ccccccccc|c}
  \hline
  \centering Datasets & $\theta_{intra}$ $\theta_{cross}$ & 10\% & 20\% & 30\% & 40\% & 50\% & 60\% & 70\% & 80\% & 90\% & Ave. mAP \\
  \hline
  \hline
  \centering \multirow{9}*{MS-COCO} & 0.1 & 64.3 & 70.1 & 73.7 & 74.7 & 74.9 & 75.7 & 76.9 & 77.8 & 78.3 & 74.0 \\
  \centering ~ & 0.2 & 67.0 & 70.2 & 73.5 & 74.2 & 74.6 & 75.5 & 76.5 & 77.7 & 79.0 & 74.2 \\
  \centering ~ & 0.3 & 63.7 & 70.5 & 72.7 & 74.4 & 75.5 & 76.0 & 76.9 & 77.8 & 79.3 & 74.1 \\
  \centering ~ & 0.4 & 67.9 & 69.5 & 72.5 & 73.9 & 74.8 & 76.0 & 78.1 & 79.2 & 80.1 & 74.6 \\
  \centering ~ & 0.5 & 64.8 & 71.5 & 72.6 & 75.6 & 77.5 & 78.5 & 79.0 & 80.2 & \textbf{80.5} & 75.6 \\
  \centering ~ & 0.6 & 67.3 & 73.8 & 76.6 & 77.1 & 78.0 & 78.4 & 79.3 & 80.0 & 80.4 & 76.8 \\
  \centering ~ & 0.7 & 70.0 & 74.3 & 76.9 & 78.0 & \textbf{79.0} & 79.3 & 79.9 & 80.2 & 80.3 & 77.5 \\
  \centering ~ & 0.8 & 68.7 & 75.2 & 77.0 & 78.2 & 78.8 & 79.3 & 79.8 & 80.2 & 80.3 & 77.5 \\
  \centering ~ & 0.9 & 67.8 & 74.8 & 76.4 & 78.2 & 78.4 & 79.0 & 79.6 & 80.1 & 80.2 & 77.2 \\
  \hline
  \centering ~ & DTL & \textbf{70.6} & \textbf{75.8} & \textbf{77.3} & \textbf{78.3} & \textbf{79.0} & \textbf{79.4} & \textbf{79.9} & \textbf{80.3} & 80.4 & \textbf{77.9} \\
  \hline
  \hline
  \centering \multirow{9}*{VG-200} & 0.1 & 35.8 & 38.3 & 39.2 & 39.8 & 40.4 & 41.6 & 42.7 & 44.0 & 45.3 & 40.8 \\
  \centering ~ & 0.2 & 36.2 & 38.4 & 39.1 & 40.9 & 41.0 & 42.2 & 42.8 & 43.6 & 44.7 & 41.0 \\
  \centering ~ & 0.3 & 36.3 & 39.1 & 39.3 & 41.3 & 43.0 & 44.0 & 44.6 & 45.0 & 45.6 & 42.0 \\
  \centering ~ & 0.4 & 39.1 & 40.9 & 41.2 & 42.2 & 43.1 & 43.9 & 44.3 & 44.9 & 45.6 & 42.8 \\
  \centering ~ & 0.5 & 39.7 & 41.0 & 42.0 & 42.6 & 43.7 & 44.9 & 46.0 & 47.0 & 47.5 & 43.8 \\
  \centering ~ & 0.6 & 39.1 & 41.4 & 44.2 & 44.5 & 45.0 & 45.4 & 46.0 & 46.3 & 46.8 & 44.3 \\
  \centering ~ & 0.7 & 39.2 & 40.6 & 43.1 & 44.4 & 44.4 & 44.8 & 45.3 & 46.3 & 47.3 & 43.9 \\
  \centering ~ & 0.8 & 38.7 & 40.6 & 43.0 & 43.8 & 44.3 & 45.4 & 45.8 & 46.1 & 46.7 & 43.8 \\
  \centering ~ & 0.9 & 39.6 & 40.5 & 42.7 & 43.9 & 44.5 & 44.9 & 45.3 & 45.8 & 46.1 & 43.7 \\
  \hline
  \centering ~ & DTL & \textbf{40.6} & \textbf{41.6} & \textbf{43.3} & \textbf{44.6} & \textbf{45.2} & \textbf{45.8} & \textbf{46.8} & \textbf{47.2} & \textbf{47.8} & \textbf{44.8} \\
  \hline
  \hline
  \centering \multirow{9}*{VOC 2007} & 0.1 & 81.5 & 84.2 & 88.6 & 89.3 & 90.2 & 91.3 & 91.9 & 92.1 & 92.3 & 89.0 \\
  \centering ~ & 0.2 & 81.9 & 85.6 & 89.3 & 90.0 & 90.7 & 91.4 & 91.7 & 92.2 & 92.4 & 89.5 \\
  \centering ~ & 0.3 & 82.4 & 87.1 & 89.5 & 90.5 & 91.4 & 91.9 & 92.0 & 92.2 & 92.3 & 89.9 \\
  \centering ~ & 0.4 & 82.8 & 87.3 & 89.8 & 91.1 & 91.4 & 92.0 & 92.2 & 92.3 & 92.5 & 90.2 \\
  \centering ~ & 0.5 & 83.3 & 87.9 & 90.2 & 91.0 & 91.6 & 92.0 & 92.2 & 92.5 & 92.5 & 90.4 \\
  \centering ~ & 0.6 & 84.0 & 89.0 & 90.2 & 91.4 & 91.9 & 92.2 & 92.3 & 92.4 & 92.6 & 90.7 \\
  \centering ~ & 0.7 & 83.0 & 88.4 & 90.9 & 91.4 & 91.7 & 91.8 & 92.4 & 92.7 & 92.6 & 90.5 \\
  \centering ~ & 0.8 & 83.1 & 87.8 & 90.3 & 91.0 & 91.7 & 92.2 & 92.2 & 92.6 & 92.6 & 90.4 \\
  \centering ~ & 0.9 & 83.5 & 88.0 & \textbf{90.5} & 91.4 & 91.7 & 92.1 & 92.4 & 92.4 & 92.5 & 90.5 \\
  \hline
  \centering ~ & DTL & \textbf{84.3} & \textbf{89.1} & \textbf{90.5} & \textbf{91.6} & \textbf{92.1} & \textbf{92.4} & \textbf{92.5} & \textbf{92.8} & \textbf{92.8} & \textbf{90.9} \\
  \hline
  \end{tabular}
  \vspace{0pt}
  \caption{Comparison of the mAPs achieved by our framework with different fixed threshold values and adaptively learned values. The best results are highlighted in bold.}
  \label{tab:fixed-threshold}
\end{table*}

\subsubsection{Analysis of SARL}
\label{sec:exp-sarl}
SARL is a basic module of the HST framework. In this work, we employ two algorithms, namely SD and SA, to implement this module. Here, we conduct experimetns to evaluate the impact of using these two implementation algorithms. As demonstrated in Tables \ref{tab:average-ablation-results-sarl}, we observe that both algorithms yield comparable performance levels. This finding indicates that the HST framework is versatile in its use of different SARL implementation algorithms and is not exclusively dependent on the SD module. Notably, when utilizing the SD algorithm (referred to as "Ours"), our model achieves average mAP values of 77.9\%, 44.8\%, and 90.9\% on the MS-COCO, VG-200, and Pascal VOC 2007 datasets, respectively. These figures slightly surpass the results obtained using the SA algorithm. Based on these outcomes, we have chosen to implement the SARL module using the SD algorithm throughout this work.

As discussed above, SARL learns category-specific features by introducing semantic information (i.e., semantic vectors and class activation maps in work \cite{chen2019learning} and \cite{Ye2020ADD-GCN}) as guidance and making predictions mainly based on the corresponding feature vectors. Here, we further visualize some examples of the input image and the semantic maps generated by SD or SA of the corresponding existing categories in Figure \ref{fig:visualization}. These results show that both SD and SA are able to highlight the semantic regions well when the objects of the corresponding categories exist.

\subsubsection{Analysis of DTL}
\label{sec:exp-dtl}
In the IST and CST modules, $\theta_{intra}$ and $\theta_{cross}$ are two crucial hyperparameters that control the precision and recall of the generated pseudo labels. We conduct experiments that vary $\theta_{intra}$ and $\theta_{cross}$ from 0.1 to 0.9 and present the results in Table \ref{tab:fixed-threshold}. We find that the optimal values for different datasets and settings are different. On one hand, for the same MS-COCO dataset with different known label proportions, the optimal thresholds the best thresholds are 0.7, 0.8, and 0.5 for the 10\%, 20\%, and 60\% known label proportion settings. On the other hand, for the same 10\% known label proportions on different datasets, the best thresholds are 0.7, 0.5, and 0.6 on the MS-COCO, VG-200, and VOC 2007 datasets. However, searching the optimal thresholds from the space \{0.1, 0.2, \dots, 0.9\} requires nearly 10 times more resources when setting $\theta_{intra}$ and $\theta_{cross}$ to the same value and nearly 100 times when setting them to different values. To avoid wasting time and resources, we propose the DTL algorithm to adaptively learn the optimal value. As shown in Table \ref{tab:fixed-threshold}, the DTL algorithm can always obtain the best mAPs under different known label proportion settings. Compared with setting them to 0.7 which achieves the overall best performance, it improves the average mAPs by 0.4\%, 0.9\%, and 0.4\% without any manual tuning.

\subsection{Analyses of Complementary Augmentation}
HST introduces and leverages semantic correlations to retrieve unknown labels, thereby enhancing MLR-PL performance. Contemporary studies \cite{pu2022semantic} also employ representation mixup techniques to generate blended features with combined labels (known as SARB). This approach is somewhat complementary with HST and has the potential to further augment its performance. In this context, we conducted an experiment integrating the feature blending strategy into HST, referred to as HST w/ FB. As depicted in Table \ref{tab:hst++}, HST demonstrates competitive performance in comparison to feature blending works, achieving identical mAP and slightly better CF1 and OF1 scores. The integration of feature blending into HST can still lead to improvement across all performance metrics, with increases in average mAP, CF1, and OF1 by 0.8\%, 0.5\%, and 0.3\%, respectively. These results indicate that HST can effectively incorporate complementary strategies to further enhance its efficacy.

\begin{table}[htb]
  \centering
  \scriptsize
  \begin{tabular}{c|ccc}
  \hline
  \centering  Methods & \multicolumn{3}{c}{Average Results} \\
  \cline{2-4}
  \centering ~ & mAP & OF1 & CF1 \\
  \hline
  \hline
  \centering   SARB & 77.9 & 76.5 & 72.2 \\
  \centering  HST &77.9 & 76.7 & 72.6 \\
  \centering  HST w/ FB & \textbf{78.7} & \textbf{77.1} & \textbf{72.9} \\
  \hline
  \end{tabular}
  \vspace{2pt}
  \caption{The average mAP, OF1 and CF1 values of SARB, HST, and HST integrated with feature blending (denoted as HST w/ FB) on the MS-COCO dataset. The best results are highlighted in bold.}
  \label{tab:hst++}
\end{table}


\section{Conclusion}
\label{sec:conclusion}
In this work, we aim to address the task of cost-efficient MLR-PL, in which only a few labels are known while the rest are unknown for each image. To address this task, we propose a novel heterogeneous semantic transfer framework that explores intra-image and cross-image semantic correlations to help complement the unknown labels. Concretely, the proposed approach consists of an IST module to learn a image-specific co-occurrence matrix for each image and transfer the semantics of known labels to the complement unknown labels and a CST module to learn category-specific feature-prototype similarities to transfer the semantics of known labels to complement the unknown labels. Finally, we conduct extensive experiments on various multi-label datasets (e.g., MS-COCO, VG-200, and Pascal VOC 2007) to demonstrate the superiority of the developed method.

\section*{Acknowledgment}
This work was supported in part by National Natural Science Foundation of China (NSFC) under Grant No. 62206060, 62272494, and 62373112, GuangDong Basic and Applied Basic Research Foundation under Grant No. SL2022A04J01626, 2023A1515012845, and 2023A1515011374, in part by Guangzhou Key R\&D Program (no. 202206010104), in part by National Key R\&D Program of China (no. 2021ZD0111600) and Natural Science Foundation of Guangdong Province (no. 2021A1515011141)

\noindent{\textbf{Code availability.} } \quad All trained models and codes are publicly available on GitHub: \url{https://github.com/HCPLab-SYSU/HCP-MLR-PL}.

\noindent{\textbf{Data availability.} } \quad The data that support the finding of this study are openly available at the following URL: \url{https://cocodataset.org}, \url{http://host.robots.ox.ac.uk/pascal/VOC}, \url{https://visualgenome.org}.


\bibliographystyle{spmpsci}      
\bibliography{referencce} 

\end{document}


\title{Heterogeneous Semantic Transfer for Multi-label Recognition with Partial Labels
}
\subtitle{Supplementary Material}


\author{Tianshui Chen \and
        Tao Pu \and
        Lingbo Liu \and
        Yukai Shi \and
        Zhijing Yang \and
        Liang Lin
}


\date{Received: date / Accepted: date}

\maketitle


\def\eg{\emph{e.g.}}
\def\etal{\emph{et al}}

\newcommand{\EQREF}{Eq.~\eqref}
\newcommand{\EQSREF}{Eqs.~\eqref}
\newcommand{\FIGREF}{Fig.~\ref}
\def\proposed{VB} 
\def\fixcolor{black}
\def\hstate{\bm {\tilde s}}
\def\rstate{\bm s}
\def\jstate{\bm s^{jn}}
\def\jpolicy{\overrightarrow{\pi}}
\def\vpref{v_{\text {pref}}}
\def\vector#1{\mbox{\boldmath $#1$}}
\def\sup#1{^{(\rm #1)}}
\def\sub#1{_{\rm #1}}
\def\supi#1{^{(#1)}}
\def\vct#1{\mbox{\boldmath $#1$}}
\def\eg{{\it e.g.}}
\def\cf{{\it c.f.}}
\def\ie{{\it i.e.}}
\def\etal{{\it et al. }}
\def\etc{{\it etc}}
\newcommand{\argmax}{\mathop{\rm argmax}\limits}
\newcommand{\argmin}{\mathop{\rm argmin}\limits}

\def\Rerr{\Delta \bm r}
\def\Terr{\Delta \bm t}
\def\Xerr{\Delta \bm x}
\def\XerrRel{\Delta \bm {\tilde x}}
\def\Xgt{\dot{\bm x}}
\def\Rgt{\dot{R}}
\def\Tgt{\dot{\bm t}}
\def\arraystretchlen{1.0}

\def\cam{c}
\def\image{\mathcal I}
\def\traj{\mathcal X}
\def\btraj{\mathcal {\bm X}}
\def\keypoints{\mathcal P}
\def\states{\mathcal S}
\def\bstates{\mathcal {\bm S}}
\def\state{\bm s}
\def\ped{\bm x}
\def\pedi{\bm p} 
\def\obs{\bm z}

\def\Fi{\bm F_r}
\def\Fp{\bm F_p}
\def\vpref{\bm w}
\def\ENERGY{{\mathcal E}}

\def\DIFF#1{\textcolor{black}{#1}}
\def\DIFFCR#1{\textcolor{black}{#1}}

\section{SARL Implementation Details}
In this work, we exploit semantic decoupling (SD) \cite{chen2019learning} and semantic attention (SA) \cite{Ye2020ADD-GCN} algorithms to implement the SARL module. Here, we further introduce the implementation details of both algorithms for the self-contained understanding of the proposed HST framework. 

\subsection{Semantic Decoupling} 
As claimed in \cite{chen2019learning}, the semantic decoupling module guide model to learn semantic-specific feature representation by introducing the category semantics. For better understanding, we comprehensively introduce the implementation details of the semantic decoupling module as follows.

As the backbone network extracting the global feature map $\textbf{f}^n$ for the input image $I^n$, the semantic decoupling module extracts a $d_s$-dimensional semantic-embedding vector using the pre-trained GloVe \cite{pennington2014glove} model for each category $c$
\begin{center}
  \begin{equation}
    \mathbf{x}_c=f_{g}(w_c),
  \end{equation}
\end{center}
where $w_c$ is the semantic word of category $c$. Then, this module incorporates the semantic vector $\mathbf{x}_c$ to guide capturing semantic-aware regions and thus learning a semantic feature representation corresponding to this category. More specifically, for each location $(w, h)$, the semantic decoupling module fuse the corresponding image feature $\mathbf{f}^n_{wh}$ and $\mathbf{x}_c$ using a low-rank bilinear pooling method \cite{kim2016hadamard}
\begin{equation}
  \tilde{\mathbf{f}}^n_{c,wh}=\mathbf{P}^T\left(\tanh\left((\mathbf{U}^T\mathbf{f}^n_{wh})\odot(\mathbf{V}^T\mathbf{x}_c)\right)\right)+\mathbf{b},
\end{equation}
where $\tanh(\cdot)$ is the hyperbolic tangent function, $\mathbf{U}\in \mathcal{R}^{N\times d_1}$, $\mathbf{V}\in \mathcal{R}^{d_s\times d_1}$, $\mathbf{P}\in \mathcal{R}^{d_1\times d_2}$, $\mathbf{b}\in \mathcal{R}^{d_2}$ are the learnable parameters, and $\odot$ is the element-wise multiplication operation. $d_1$ and $d_2$ are the dimensions of the joint embeddings and the output features. Then, an attentional coefficient is computed under the guidance of $\mathbf{x}_c$ by
\begin{equation}
  \tilde{a}_{c,wh}=f_{a}(\tilde{\mathbf{f}}^n_{c,wh}).
\end{equation}
where $f_{a}(\cdot)$ is an attentional function and it is implemented by a fully connected network. The process is repeated for all locations. To make the coefficients easily comparable across different samples, we normalize the coefficients over all locations using a softmax function
\begin{equation}
  a_{c,wh}=\frac{\mathrm{exp}(\tilde{a}_{c,wh})}{\sum_{w',h'}{\mathrm{exp}(\tilde{a}_{c,w'h'}})}.
\end{equation}
Finally, we perform weighted average pooling over all locations to obtain a feature vector
\begin{equation}
  \mathbf{f}^n_{c}=\sum_{w,h}a_{c,wh}\mathbf{f}^n_{c,wh},
\end{equation}
that encodes the information related to category $c$. We repeat the process for all categories and obtain a feature vector for each category $[\mathbf{f}^n_{1}, \mathbf{f}^n_{2}, \dots, \mathbf{f}^n_{C}]$.

\subsection{Semantic Attention Module}
As designed in \cite{Ye2020ADD-GCN}, the objective of the semantic attention module is to learn content-aware category representations that focus on the content regions for each category. For depth of understanding, we comprehensively introduce the implementation details of the semantic attention module as follows.

\begin{table*}
  \centering
  \small
  \begin{tabular}{c|c|ccccccccc|c}
  \hline
  \centering Datasets & Methods & 10\% & 20\% & 30\% & 40\% & 50\% & 60\% & 70\% & 80\% & 90\% & Ave. OF1 \\
  \hline
  \hline
  \centering \multirow{8}*{MS-COCO} & SSGRL & 64.7 & 71.6 & 72.9 & 74.3 & 75.2 & 76.0 & 76.3 & 76.9 & 77.2 & 73.9 \\
  \centering ~ & GCN-ML & 65.5 & 68.6 & 71.9 & 73.2 & 75.3 & 75.5 & 75.7 & 75.8 & 76.4 & 73.1 \\
  \centering ~ & KGGR & 66.7 & 69.4 & 72.8 & 75.1 & 75.1 & 75.3 & 75.8 & 76.2 & 76.6 & 73.7 \\
  \centering ~ & P-GCN & - & - & - & - & - & - & - & - & - & - \\
  \centering ~ & ASL & 31.7 & 41.3 & 46.3 & 47.2 & 46.8 & 51.0 & 51.5 & 51.3 & 53.0 & 46.7 \\
  \centering ~ & CL & 29.0 & 33.8 & 55.0 & 66.6 & 72.3 & 72.4 & 73.8 & 76.9 & 77.1 & 61.9 \\
  \centering ~ & Partial BCE & 64.0 & 70.9 & 73.6 & 75.3 & 75.9 & 76.4 & 76.7 & 76.8 & 76.8 & 74.0 \\
  \centering ~ & HST & \textbf{72.1} & \textbf{75.1} & \textbf{76.2} & \textbf{76.8} & \textbf{77.3} & \textbf{77.8} & \textbf{78.0} & \textbf{78.4} & \textbf{78.6} & \textbf{76.7} \\
  \hline
  \hline
  \centering \multirow{8}*{VG-200} & SSGRL & 33.0 & 34.9 & 37.1 & 37.7 & 38.4 & 39.0 & 39.8 & 40.0 & 40.1 & 37.8 \\
  \centering ~ & GCN-ML & 28.3 & 36.2 & 38.1 & 39.6 & 39.8 & 40.5 & 40.9 & 42.1 & 42.5 & 38.7 \\
  \centering ~ & KGGR & 38.2 & 39.8 & 40.7 & 41.0 & 41.8 & 41.9 & 41.9 & 42.5 & 42.7 & 41.2 \\
  \centering ~ & P-GCN & - & - & - & - & - & - & - & - & - & - \\
  \centering ~ & ASL & 22.3 & 24.9 & 24.8 & 25.3 & 25.2 & 25.5 & 25.8 & 25.9 & 24.5 & 24.9 \\
  \centering ~ & CL & 9.2 & 14.8 & 20.4 & 22.1 & 23.4 & 26.6 & 29.9 & 32.2 & 33.6 & 23.6 \\
  \centering ~ & Partial BCE & 20.8 & 33.7 & 35.4 & 36.8 & 37.6 & 39.4 & 39.7 & 40.7 & 40.9 & 36.1 \\
  \centering ~ & HST & \textbf{43.1} & \textbf{43.6} & \textbf{45.6} & \textbf{46.0} & \textbf{46.4} & \textbf{47.7} & \textbf{47.7} & \textbf{48.3} & \textbf{48.2} & \textbf{46.3} \\
  \hline
  \hline
  \centering \multirow{8}*{VOC2007} & SSGRL & 80.6 & 86.5 & 87.9 & 88.3 & 88.7 & 89.1 & 89.2 & 89.4 & 89.3 & 87.7 \\
  \centering ~ & GCN-ML & 82.7 & 84.7 & 87.2 & 87.9 & 88.1 & 88.2 & 88.9 & 89.0 & 89.2 & 87.3 \\
  \centering ~ & KGGR & 78.7 & 84.9 & 85.4 & 86.6 & 88.5 & 88.5 & 88.6 & 88.6 & 88.7 & 86.5 \\
  \centering ~ & P-GCN & - & - & - & - & - & - & - & - & - & - \\
  \centering ~ & ASL & 30.7 & 33.0 & 36.1 & 39.2 & 43.8 & 46.0 & 45.9 & 47.3 & 47.3 & 41.0 \\
  \centering ~ & CL & 53.9 & 81.1 & 87.1 & 87.9 & 88.2 & 88.9 & 88.9 & 89.2 & 89.3 & 83.8 \\
  \centering ~ & Partial BCE & 82.0 & 86.5 & 87.7 & 88.1 & 88.6 & 89.2 & 89.7 & 89.7 & 89.7 & 87.9 \\
  \centering ~ & HST & \textbf{83.6} & \textbf{87.1} & \textbf{87.9} & \textbf{88.9} & \textbf{89.3} & \textbf{89.5} & \textbf{90.0} & \textbf{89.9} & \textbf{89.8} & \textbf{88.4} \\
  \hline
  \end{tabular}
  \vspace{0pt}
  \caption{The average OF1 and OF1 values achieved by our HST framework and current state-of-the-art competitors with different known label proportions for MLR-PL on the MS-COCO, VG-200, and Pascal VOC 2007 datasets. The best results are highlighted in bold.}
  \vspace{-10pt}
  \label{tab:OF1-results}
\end{table*}

Given the input feature map $\mathbf{f}^n\in\mathbb{R}^{{D}\times{W}\times{H}}$, the semantic attention module first calculates category-specific activation maps based on Class Activation Mapping (CAM)\cite{Zhou2016CAM}
\begin{equation}
\mathbf{M}=[\mathbf{m}_1,\mathbf{m}_2,\dots,\mathbf{m}_C] \in \mathbb{R}^{{C}\times{W}\times{H}},
\end{equation}
where $D, W, H$ are the channel dimension, width and height of the global feature map $\mathbf{f}^n$ and $C$ is the category number. Specifically, we can perform Global Average Pooling (GAP) or Global Max Pooling (GMP) on the feature map $\mathbf{f}^n$ and classify these pooled features with FC classifiers. Then these classifiers are used to identify the category-specific activation maps by convolving the weights of FC classifiers with feature map $\mathbf{f}^n$.

Based on these category-specific activation maps $\mathbf{M}$, the semantic attention module then convert the transformed feature map $\tilde{\mathbf{f}}^n \in \mathbb{R}^{D' \times W \times H}$ to generate the content-aware category representations $[\mathbf{f}^n_1,\mathbf{f}^n_2,\dots,\mathbf{f}^n_C] \in \mathbb{R}^{C \times D}$. Specifically, each class representation $\mathbf{f}^n_c$ is formulated as a weighted sum on $\tilde{\mathbf{f}}^n$ as follows, such that the produced $\mathbf{f}^n_c$ can selectively aggregate features related to its specific category $c$.
\begin{equation}
    \mathbf{f}^n_c=\mathbf{m}_c^T\tilde{\mathbf{f}}^n=\sum_{i=1}^{H}\sum_{j=1}^{W}{m_{i,j}^{c} \tilde{\mathbf{f}}^n_{i,j}},
\end{equation}
where $m_{i,j}^{c}$ and $\tilde{\mathbf{f}}^n_{i,j} \in \mathbb{R}^{D'}$ are the weight of \textit{c}~th activation map and the feature vector of the feature map at $(i,j)$, respectively. 

\begin{table*}
  \centering
  \small
  \begin{tabular}{c|c|ccccccccc|c}
  \hline
  \centering Datasets & Methods & 10\% & 20\% & 30\% & 40\% & 50\% & 60\% & 70\% & 80\% & 90\% & Ave. CF1 \\
  \hline
  \hline
  \centering \multirow{8}*{MS-COCO} & SSGRL & 54.3 & 64.3 & 66.9 & 68.2 & 70.3 & 71.5 & 71.7 & 72.6 & 73.0 & 68.1 \\
  \centering ~ & GCN-ML & 58.1 & 62.4 & 66.6 & 68.5 & 71.5 & 71.6 & 71.8 & 72.3 & 72.5 & 68.4 \\
  \centering ~ & KGGR & 62.6 & 64.8 & 68.8 & 70.7 & 71.5 & 71.7 & 72.1 & 72.5 & 73.0 & 69.7 \\
  \centering ~ & P-GCN & - & - & - & - & - & - & - & - & - & - \\
  \centering ~ & ASL & 29.6 & 41.2 & 46.9 & 48.4 & 49.2 & 52.6 & 53.8 & 53.7 & 56.1 & 47.9 \\
  \centering ~ & CL & 3.7 & 3.7 & 32.4 & 54.4 & 64.3 & 64.4 & 66.8 & 72.4 & 72.9 & 48.3 \\
  \centering ~ & Partial BCE & 54.6 & 64.2 & 68.5 & 70.6 & 71.5 & 72.2 & 72.4 & 72.6 & 72.6 & 68.8 \\
  \centering ~ & HST & \textbf{66.4} & \textbf{70.7} & \textbf{72.1} & \textbf{72.8} & \textbf{73.4} & \textbf{74.0} & \textbf{74.1} & \textbf{74.8} & \textbf{75.0} & \textbf{72.6} \\
  \hline
  \hline
  \centering \multirow{8}*{VG-200} & SSGRL & 20.0 & 22.6 & 25.3 & 26.5 & 26.5 & 27.4 & 28.7 & 28.7 & 28.9 & 26.1 \\
  \centering ~ & GCN-ML & 14.4 & 18.8 & 23.2 & 24.1 & 27.1 & 29.4 & 30.4 & 31.3 & 32.0 & 25.6 \\
  \centering ~ & KGGR & 25.3 & 28.9 & 30.9 & 34.2 & 35.2 & 35.7 & 36.0 & 37.1 & 38.9 & 33.6 \\
  \centering ~ & P-GCN & - & - & - & - & - & - & - & - & - & - \\
  \centering ~ & ASL & 21.0 & 23.4 & 23.3 & 23.5 & 23.6 & 23.6 & 24.3 & 23.9 & 22.8 & 23.3 \\
  \centering ~ & CL & 2.1 & 3.3 & 6.8 & 7.3 & 9.1 & 11.9 & 16.9 & 19.3 & 21.3 & 10.9 \\
  \centering ~ & Partial BCE & 6.9 & 23.0 & 25.7 & 27.2 & 27.7 & 29.5 & 29.6 & 30.2 & 31.1 & 25.7 \\
  \centering ~ & HST & \textbf{30.3} & \textbf{34.5} & \textbf{37.2} & \textbf{38.4} & \textbf{38.6} & \textbf{40.1} & \textbf{40.6} & \textbf{40.7} & \textbf{40.9} & \textbf{37.9} \\
  \hline
  \hline
  \centering \multirow{8}*{VOC2007} & SSGRL & 73.9 & 82.9 & 85.0 & 85.6 & 86.2 & 86.4 & 86.6 & 86.7 & 87.0 & 84.5 \\
  \centering ~ & GCN-ML & 78.2 & 81.2 & 84.5 & 85.4 & 85.7 & 86.0 & 86.6 & 86.8 & 86.9 & 84.6 \\
  \centering ~ & KGGR & 75.2 & 82.5 & 84.4 & 85.3 & 86.9 & 87.0 & 87.0 & 87.0 & 87.0 & 84.7 \\
  \centering ~ & P-GCN & - & - & - & - & - & - & - & - & - & - \\
  \centering ~ & ASL & 31.1 & 34.0 & 36.0 & 39.7 & 43.5 & 45.5 & 45.7 & 46.3 & 46.4 & 40.9 \\
  \centering ~ & CL & 13.5 & 64.5 & 83.4 & 85.2 & 85.6 & 86.3 & 86.3 & 86.8 & 86.8 & 75.4 \\
  \centering ~ & Partial BCE & 76.1 & 83.2 & 84.4 & 85.0 & 86.0 & 86.7 & 87.2 & 87.3 & 87.5 & 84.8 \\
  \centering ~ & HST & \textbf{79.2} & \textbf{84.7} & \textbf{85.3} & \textbf{86.9} & \textbf{87.2} & \textbf{87.7} & \textbf{88.1} & \textbf{88.1} & \textbf{87.8} & \textbf{86.1} \\
  \hline
  \end{tabular}
  \vspace{0pt}
  \caption{The average CF1 and CF1 values achieved by our HST framework and current state-of-the-art competitors with different known label proportions for MLR-PL on the MS-COCO, VG-200, and Pascal VOC 2007 datasets. The best results are highlighted in bold.}
  \vspace{-10pt}
  \label{tab:CF1-results}
\end{table*}

\redm{\section{Evaluation Metrics}}
\redm{In the original paper, we utilize the mean Average Precision (mAP) metric for evaluating performance across various proportions of known labels. This process involves arranging the prediction results in descending order based on their respective scores, followed by calculating the Average Precision (AP) for each category. The mean Average Precision (mAP) is then determined by averaging these AP values across all categories, providing a comprehensive measure of the overall performance. mAP is the most widely used metric across variance MLR works, and we mainly report it in the original paper.}

\redm{To enhance the thoroughness of our evaluations, we also present the Overall F1 (OF1) and Per-Class F1 (CF1) metrics, which are commonly employed in assessing MLR models \cite{chen2019learning,chen2019multi,chen2021p-gcn}. To calculate OF1 and CF1, it's essential first to determine the Overall Precision and Recall (OP, OR) as well as the Per-Class Precision and Recall (CP, CR). In our discussion, we provide detailed definitions of OP, OR, OF1, CP, CR, and CF1. These metrics can be formally computed by
\begin{gather}
 OP=\frac{\sum_{i}{N^c_i}}{\sum_{i}{N^p_i}}, CP=\frac{1}{C} \sum_{i}{ \frac{N^c_i}{N^p_i} } \\
 OR=\frac{\sum_{i}{N^c_i}}{\sum_{i}{N^g_i}}, CR=\frac{1}{C} \sum_{i}{ \frac{N^c_i}{N^g_i} } \\
 OF1=\frac{2 \times OP \times OR }{OP+OR},  CF1=\frac{2 \times CP \times CR }{CP+CR}
\end{gather}
where $N^c_i$ is the number of images that are correctly predicted for the $i$-th label, $N^p_i$ is the number of predicted images for the $i$-th label, $N^g_i$ is the number of ground truth images for the $i$-th label. We also average the OF1 and CF1 over all known label proportions. }

\section{Supplementary Experiments}
Due to the page limit, we merely present the average OF1 and CF1 in the original manuscript. Here, we further present the OF1 and CF1 on different known label proportion settings. As shown in Table \ref{tab:OF1-results} and \ref{tab:CF1-results}, HST still obtains the best OF1 and CF1 on all known label proportion settings.




\bibliographystyle{spmpsci}      
\bibliography{referencce} 